\documentclass[letterpaper, 10 pt, journal, twoside]{IEEEtran}
\IEEEoverridecommandlockouts                              % This command is only needed if 

\usepackage{hyperref}
\hypersetup{
    colorlinks=true,
    linkcolor=blue,
    filecolor=magenta,      
    urlcolor=cyan,
}
 
\usepackage{subfigure}
\usepackage{mathtools} % loads amsmath 
\usepackage{graphicx}
\usepackage{wrapfig}
\usepackage{multicol}
\usepackage{booktabs}
\usepackage{amsfonts}
\usepackage{gensymb}
\usepackage{listings}
\usepackage{multirow}
\usepackage{pifont}
\usepackage{dsfont}

\usepackage{algorithm}
\usepackage{algpseudocode}
\usepackage{epstopdf}
\usepackage{commath}

\usepackage{array,tabularx,calc}

\usepackage{makecell}
\usepackage{colortbl}
\definecolor{light-gray}{gray}{0.9}
\newcolumntype{C}{>{\columncolor{light-gray}}c}
\newcolumntype{L}{>{\columncolor{light-gray}}l}
\newcolumntype{R}{>{\columncolor{light-gray}}r}

\DeclareMathOperator*{\argmax}{arg\,max}

\algnewcommand{\LineComment}[1]{\State \(\triangleright\) #1}

\usepackage[font=scriptsize,labelfont=bf]{caption}
\newcommand{\1}{\mathds{1}}

\DeclareMathOperator{\progress}{{\mathcal{P}}}
\DeclareMathOperator{\rprogress}{R_{\progress}}
\DeclareMathOperator{\rtrial}{{\mathds{R}_{\text{trial}}}}
\DeclareMathOperator{\rdiscount}{{\mathds{R}_{\mathcal{D}}}}

\newcommand{\xmark}{\text{\ding{55}}}
\newcommand{\cmark}{\text{\ding{51}}}

\begin{document}
% \title{\LARGE \bf
% ``Good Robot!'':\\
% Efficient Reinforcement Learning for Multi-Step\\
% Visual Tasks via Reward Shaping}
\title{\LARGE \bf
``Good Robot!'':\\
Efficient Reinforcement Learning for Multi-Step\\
Visual Tasks with Sim to Real Transfer}
% \title{\LARGE \bf
% ``Good Robot!'':\\
% Efficient Reinforcement Learning for Multi-Step\\
% Visual Tasks via Dynamic Action Spaces}

%\author{Andrew~Hundt,~\IEEEmembership{Student Member,~IEEE,},
%        Heeyeon Kwon,
%        Chris~Paxton,~\IEEEmembership{Member,~IEEE,}
%        and~Greg~Hager,~\IEEEmembership{Fellow,~IEEE}% %<-this % stops a space
\author{Andrew Hundt$^1$, Benjamin Killeen$^1$, Nicholas Greene$^1$, Hongtao Wu$^1$,\\ Heeyeon Kwon$^1$, Chris Paxton$^2$, and Gregory D. Hager$^1$
% \thanks{Manuscript received: February, 24, 2020; Revised: June, 21, 2020; Accepted: July, 20, 2020.}%Use only for final RAL version
\thanks{Manuscript received: February, 24, 2020; Accepted: July, 20, 2020.}%Use only for final ArXiV version
\thanks{This letter was recommended for publication by Associate Editor J. Kober and Editor T. Asfour upon evaluation of the Reviewers' comments.
This work was supported by the NSF NRI Awards \#1637949 and \#1763705, and in part by Office of Naval Research Award N00014-17-1-2124. \textit{(Corresponding author: Andrew Hundt)} } %Use only for final RAL version
\thanks{$^{1}$The Johns Hopkins University. {\small \tt \{ahundt, killeen, ngreen29, hwu67, hkwon28, ghager1\} @jhu.edu}}
\thanks{$^{2}$NVIDIA. {\small \tt cpaxton@nvidia.com}}
% \thanks{Digital Object Identifier (DOI): see top of this page.} % Use only for the final RAL version
\thanks{Digital Object Identifier (DOI): \url{http://doi.org/10.1109/LRA.2020.3015448}} % Use only for final ArXiV version
}

\markboth{IEEE Robotics and Automation Letters. Preprint Version. Accepted July, 2020}
{Hundt \MakeLowercase{\textit{et al.}}: GoodRobot} 
% Use only for final RAL version

\maketitle

\begin{abstract}

Current Reinforcement Learning (RL) algorithms struggle with long-horizon tasks where time can be wasted exploring dead ends and task progress may be easily reversed. 
We develop the SPOT framework, which explores within action safety zones, learns about unsafe regions without exploring them, and prioritizes experiences that reverse earlier progress to learn with remarkable efficiency. 

The SPOT framework successfully completes simulated trials of a variety of tasks, improving a baseline trial success rate from 13\% to 100\% when stacking 4 cubes, from 13\% to 99\% when creating rows of 4 cubes, and from 84\% to 95\% when clearing toys arranged in adversarial patterns.
Efficiency with respect to actions per trial typically improves by 30\% or more, while training takes just 1-20k actions, depending on the task.

Furthermore, we demonstrate direct sim to real transfer.
We are able to create real stacks in 100\% of trials with 61\% efficiency and real rows in 100\% of trials with 59\% efficiency by directly loading the simulation-trained model on the real robot with no additional real-world fine-tuning.
To our knowledge, this is the first instance of reinforcement learning with successful sim to real transfer applied to long term multi-step tasks such as block-stacking and row-making with consideration of \textit{progress reversal}.
Code is available at \url{https://github.com/jhu-lcsr/good_robot}.

\end{abstract}

% In your text, make sure to put at least two and up to five  RAL keywords just beneath the abstract
% Keywords appear just beneath the abstract. Use only for final RAL version. 

% \fi
\IEEEpeerreviewmaketitle

\section{Introduction}

\label{sec:introduction}

\IEEEPARstart{M}{ulti-step} robotic tasks in real-world settings are notoriously challenging to learn. % Use only for the RAL version
% Multi-step robotic tasks in real-world settings are notoriously challenging to learn.
They intertwine learning the immediate physical consequences of actions with the need to understand how these consequences affect progress towards the overall goal.
Furthermore, in contrast to traditional motion planning, which assumes perfect information and known action models, learning only has access to the spatially and temporally limited information from sensing the environment.

\begin{figure}[bt!]
    \centering
    \includegraphics[width=0.9\columnwidth]{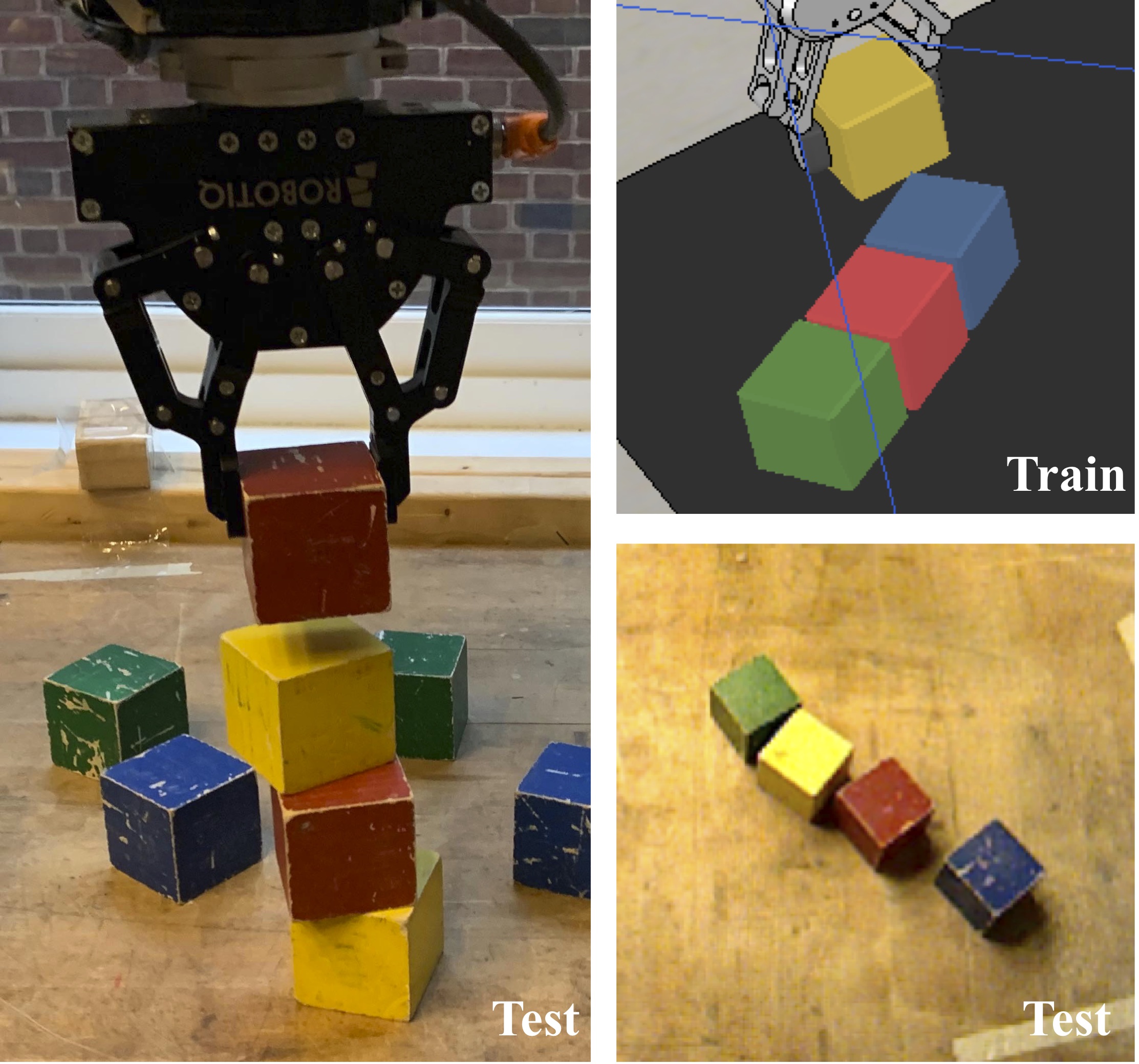}
    \caption{
    \label{fig:stacking_screenshot}
    Robot-created stacks and rows of cubes with sim to real transfer. Our Schedule for Positive Task (SPOT) framework allows us to efficiently find policies which can complete multi-step tasks.
    Video overview: \url{https://youtu.be/MbCuEZadkIw}
    % Second version, with lower final real test success rates:
    % Video overview: \url{https://youtu.be/PzU-UfzLXUM}
    % First version with EVT video overview, 3min: \url{https://youtu.be/p2iTSEJ-f_A}
}
\vspace{-0.5cm}
\end{figure}

Our key observation is that reinforcement learning wastes significant time exploring actions which are unproductive at best.
For example, in a block stacking task (Fig.~\ref{fig:stacking_screenshot}), the knowledge that grasping at empty air will never snag an object is ``common sense'' for humans, %, as is the understanding that typical robots are not safe in lava (Fig.~\ref{fig:safety_gridworld}). 
%However, a vanilla learning algorithm has to discover these principles.
but may take some time for a vanilla algorithm to discover.
To address this, we propose the Schedule for Positive Task (SPOT) framework, which 
%By contrast, the SPOT framework can
incorporates common sense constraints in a way that 
significantly accelerates both learning and final task efficiency.

%Our models focus on learning core concepts,
%for example, grasping an object from a stack does not contribute to making the stack higher; the robot should focus on lone, unstacked objects.

\begin{figure*}[bt!]
    \vspace{0.2cm}
    \centering
    % \hfill
    % \includegraphics[width=\textwidth]{StackMotionQNetwork.pdf}
    \includegraphics[width=0.9\textwidth]{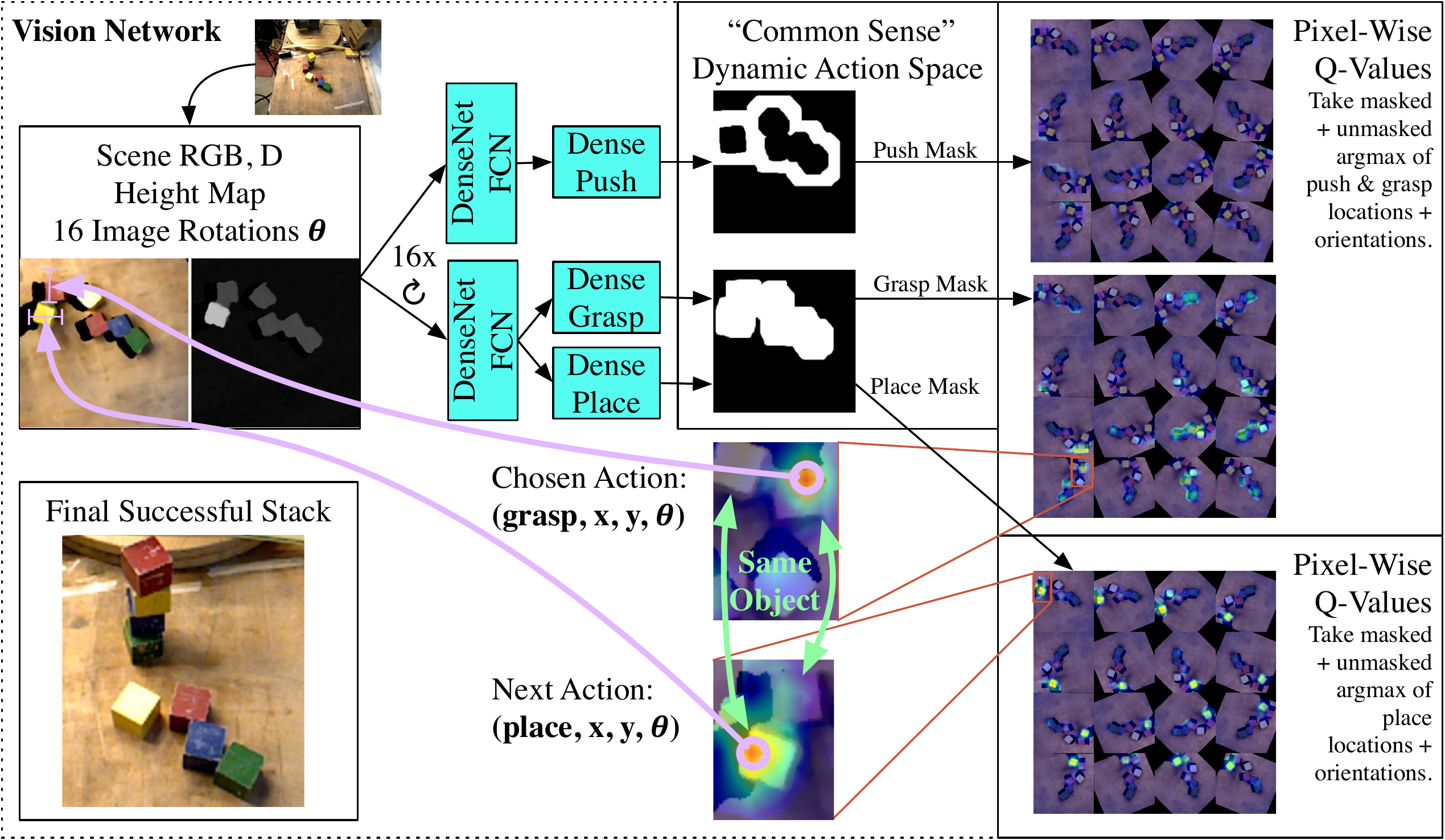} % final RAL version 
    \vspace{-0.2cm}
    \caption{
    \label{fig:stack_EVT_q_network}
    Our model architecture. 
    Images are pre-rotated to 16 orientations $\theta$ before being passed to the network. 
    Every coordinate $a = (\phi,x,y,\theta)$ in the output pixel-wise Q-Values corresponds to a final gripper position, orientation, and open loop action type, respectively.
    Purple circles highlight the highest likelihood action $\argmax_a (Q(s,M(a)))$ (Eq. \ref{eq:spotq}) with an arrow to the corresponding height map coordinate, showing how these values are transformed to a gripper pose.
    The rotated overhead views overlay the Q value at each pixel from dark blue values near 0 to red for high probabilities. 
    % Green arrows identify the same object across two oriented views; take a moment to compare the score of a single object across all actions. 
    % Each object is scored in a way which leads to a successful stack in accordance with its surrounding context. 
    If you take a moment to compare the Q values of a single object across all actions (green arrows identify the same object across two oriented views) you will see each object is scored in a way which leads to a successful stack in accordance with its surrounding context.
    For example, the grasp model learns to give a high score to the lone unstacked red block for grasp actions and a low score to the yellow top of the stack, while the place model does the reverse.
    Here the model chooses to grasp the red block and place on the yellow, blue, and green stack. 
    Experiment details are in Sec.~\ref{sec:experiments} and \ref{subsec:real_robot_experiments}.
}
% \vspace{-0.7cm}
\vspace{-0.3cm} % final RAL version
\end{figure*}
While these types of constraints are intuitive, incorporating them into Deep RL (DRL) in a manner that leads to reliable and efficient learning is nontrivial \cite{zeng2018vpg, kroemer2019a}.
%Discovering and effectively applying contextual knowledge is nontrivial, so behavior which demonstrates an understanding of context should be rewarded. 
Our methods~(Sec. \ref{sec:approach}) take inspiration from a humane and effective approach to training pets sometimes called ``Positive Conditioning.''
Consider the goal of training a dog
``Spot'' to ignore an object or event she finds particularly interesting on command.
%In this practice,
Spot is rewarded with treats whenever partial compliance with the desired end
behavior is shown, and simply removed from %harmful or
regressive situations with zero treats (reward).
One way to achieve this is to start with multiple treats in hand, place one
treat in view of Spot, and, if she eagerly jumps at the treat
(a negative action), the human snatches and hides the treat
immediately for zero reward on that action. With repetition,
Spot will eventually hesitate, and so she is immediately
praised with ``Good Spot!'' and gets a treat separate from the
one she should ignore.
This approach can be expanded to new situations and behaviors, and it encourages exploration and rapid improvement once an initial partial success is achieved.
As we describe in Sec~\ref{sec:approach}, our reward functions and SPOT-Q Learning are likewise designed to provide neither reward nor punishment
for actions that reverse progress.

Instances of \textit{progress reversal} are associated with varying complexity.
On the one hand, failing to stack the first block on top of another leaves the robot in a similar situation, so recovery takes $\Omega(1)$ actions.
However, once a stack of $n$ blocks exists, even a successful grasp might knock the whole stack down, reversing the entire history of actions for a given trial (Fig. \ref{fig:gripper_asymmetry_time_dependencies}), so recovery is $\Omega(n)$. 
The latter, more dramatic instance of \textit{progress reversal} is a challenging problem for reinforcement learning of multi-step tasks in robotics; our work provides a method for efficiently solving such cases. 

In summary, our contributions in this article are: 
(1) The overall SPOT framework for reinforcement learning of multi-step tasks, which improves on state of the art in simulation and can train efficiently on real-world situations.
(2) SPOT-Q Learning, a method for safe and efficient training in which a mask focuses exploration at runtime and generates extra on-the-fly training examples from past experience during replay.
(3) State of the art zero-shot domain transfer from simulated stacking and row building tasks to their real world counterparts, as well as robustness with respect to a change in hardware and scene positions.
(4) An ablation study showing that Situation Removal dramatically decreases \textit{progress reversal}; that a progress metric increases efficiency; and that trial rewards improve on discounting, but involve a trade-off between efficiency and support for sparse rewards.

\section{Related Work}

Deep Neural Networks (DNNs) %in particular 
have enabled the use of raw sensor data in robotic manipulation~\cite{levine2015deepvisiomotor,levine2016learning,kalashnikov2018qt,zeng2018vpg, kroemer2019a}.
In some approaches, a DNN's output directly corresponds to motor commands, \textit{e.g.} \cite{levine2015deepvisiomotor, levine2016learning}.
Higher-level methods, on the other hand, assume a simple model for robotic control and focus on bounding box or pose detection for downstream grasp planning \cite{redmon2015real,drost2010model,
Kumra_2017_resnet_grasp,levine2017semanticgrasping,cornellgrasping, 2018grasploop,zeng2018vpg}. 
% Removed citation: ,zhang2018real
RGB-D sensors can be beneficial \cite{zeng2018vpg,2018grasploop,murali20196}, as they capture physical information about the workspace. 
% However, the agent must still develop physical intuition, which recent work attempts in a more targeted setting.
Object-centric skill learning can be effective and generalize well, \textit{e.g.}~\cite{gupta2016learning,devin2018deep}. 
\cite{lerer2016learning, groth2018shapestacks} focus on block stacking by classifying simulated stacks as stable or likely to fall. 
% Of these, the ShapeStacks dataset \cite{groth2018shapestacks} includes a larger variety of objects, adding cylinders and spheres. 
Similarly, \cite{finn2016unsupervised, byravan2017se3} develop physical intuition by predicting push action outcomes. 
Our work differs by developing visual understanding and physical intuition in concert with the progress of multi-step tasks.

Grasping is a particularly active area of research. DexNet~\cite{mahler2017dex,mahler2019learning} learns from a large number of depth images of top-down grasps, and gets extremely good performance on grasping novel objects but does not look at long-horizon tasks. 6-DOF GraspNet~\cite{mousavian20196} uses simulated grasp data to generalize to new objects and has been extended to handle reliable grasping of novel objects in clutter~\cite{murali20196}.

DRL has proven effective at increasingly complex tasks in robotic manipulation~\cite{zhang2016reaching,zeng2018vpg,zeng2019tossingbot,kalashnikov2018qt}. QT-Opt~\cite{kalashnikov2018qt} learns manipulation skills from hundreds of thousands of real-world grasp attempts on real robots.
Domain Adaptation, such as applying random textures in simulation, can also enhance sim to real transfer \cite{tobin2017simtorealworld,bousmalis2018using}.
Other
methods focus on transferring visuomotor skills from simulated to real
robots \cite{zhang2016reaching, 2018visiomotordeepmind}.
Our work directs a
low-level controller to perform actions rather than regressing torque vectors
directly, following prior work \cite{zeng2018vpg, zeng2019tossingbot} by learning a pixel-wise success likelihood map. 

Multi-step tasks with sparse rewards present a particular challenge in reinforcement learning because solutions are less likely to be discovered through random exploration.
When available, demonstration can be an effective method for guiding exploration \cite{xu2018neural, aytar2018playinghardyoutube,hundt2019costar}. 
Multi-step
tasks can be split into modular sub-tasks comprising a sketch~\cite{andreas2017modular}, while \cite{devin2016modular} has 
robot-specific and task-specific learning modules. % for final RAL version
% robot and task specific learning modules.

Safety is crucial for reinforcement learning in many real-world settings~\cite{amodei2016safety,leike2017aisafety,everitt2018agisafety}. The preliminary experiments in Sec. \ref{sec:lava_gridworld_safety_domain_generalization} show that SPOT-Q provides a way to incorporate safety into general Q-Learning based algorithms~\cite{hessel2018rainbow}. 
% , and that it has useful safety applications not considered in the most comparable robotics research~\cite{zeng2018vpg,2018qtopt, zhu2018reinforcement,matas2018sim}.

We compare the SPOT framework to VPG~\cite{zeng2018vpg}, a method for RL-based table clearing tasks which can be trained from images within hours on a single robot, in Sec. \ref{sec:experiments} and \ref{subsec:real_robot_experiments}.
VPG is frequently able to complete adversarial scenarios 
% like those in Fig.~\ref{fig:grasp_efficiency},
like first pushing a tightly packed group of blocks apart and then grasping the now-separated objects.
%We utilize the VPG V-REP simulation and models as our baseline for comparison.
% VPG assumes an instantaneous reward delivery is sufficient to complete the task at hand. 
% Following VPG, Form2Fit~\cite{zakka2019form2fit} tackles kitting tasks with well defined containers within which objects should be placed, but is not investigating DRL.

\begin{figure}[bt!]
\vspace{0.2cm}
    \centering
    \includegraphics[width=0.9\columnwidth]{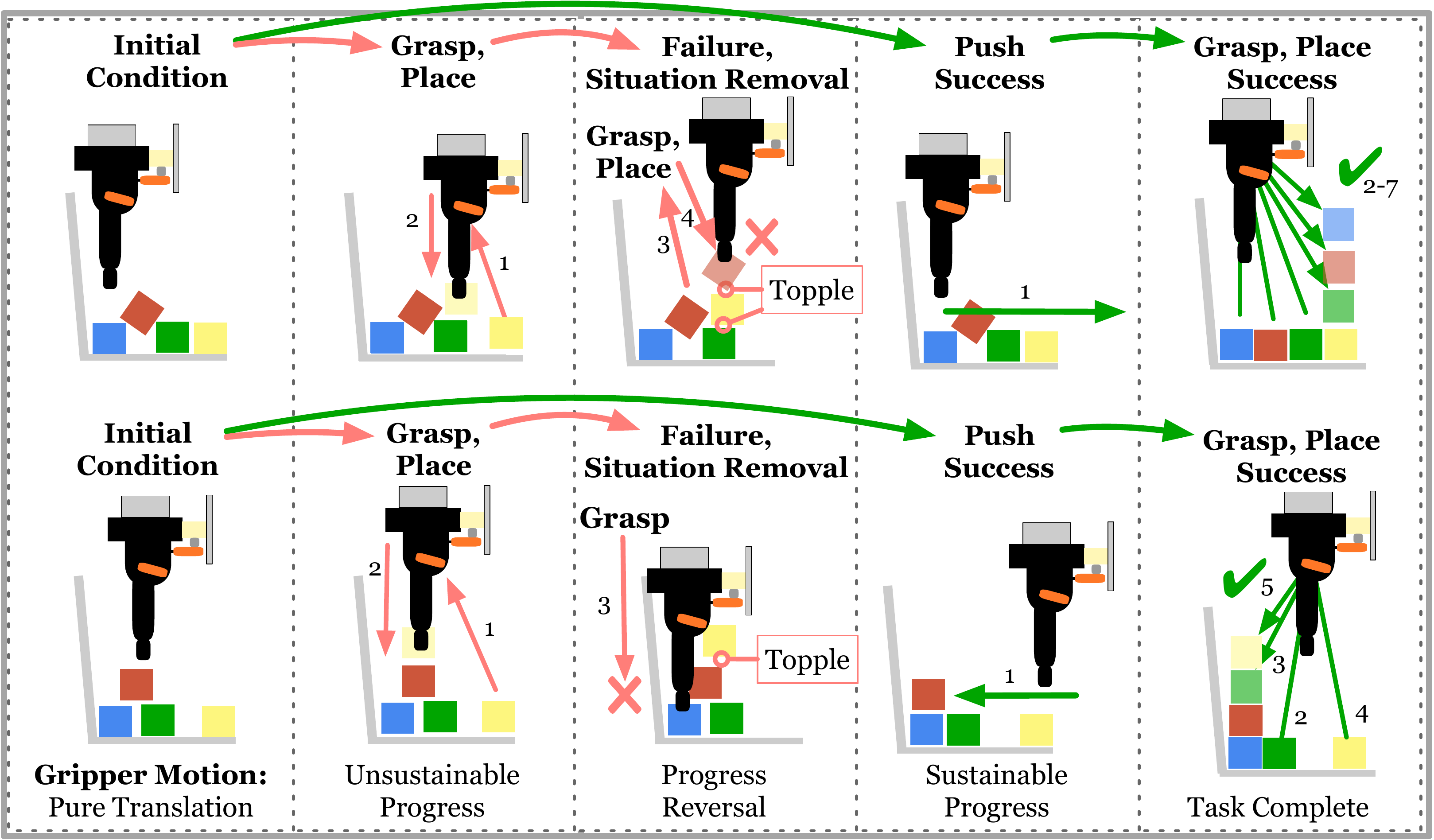}
    \caption{
    \label{fig:gripper_asymmetry_time_dependencies}
    Red arrows show how individual successful actions can fail on the larger stacking task, forcing eventual \textit{progress reversal} where a partial stack topples or the top must be removed.
    Ideally algorithms should efficiently learn to prevent this situation and succeed as indicated by the green arrows.
    % Temporal and workspace dependencies when stacking four blocks must be considered because
    % Task time dependencies can propagate forwards and backwards through time. 
    Thus, temporal and workspace dependencies must be considered. 
    Events at a current time $t_i \in T, i \in [1..n]$ can influence the likelihood of successful outcomes for past actions $t_{h} | h < i$ and future actions $t_{j} | j > i$.
    A successful choice of action at any given $t_i$ will ensure both past and future actions are productive contributors to the larger task at hand.
    % , while failures indicate either a lack of progress or regression to an earlier stage.
    In our experiments a partial stack or row is itself a scene obstacle. 
    The gray wall pictured here is for illustrative purposes only.
    % While an action will ideally enable future progress, it may instead have no effect or even undo past progress.
}
\vspace{-0.3cm}
\end{figure}

Some of the most closely related recent work involves tasks with multiple actions: \cite{zhu2018reinforcement} includes placing one block on another, \cite{matas2018sim} places one towel on a bar, and \cite{mahler2017learning} clears a bin, but the first two are not long-horizon tasks and the possibility of \textit{progress reversal} (Fig. \ref{fig:gripper_asymmetry_time_dependencies}) is never considered. %, lacunae which are among our key considerations. 

% The results video of \cite{2018visiomotordeepmind}, even includes a stack best described as a ``leaning tower of pisa'', upon which any additional blocks would lead that stack to topple.
% Similarly, in \cite{matas2018sim} they place a towel on a bar, but that work does not examine placing a second towel over the bar, which might accidentally knock the first towel off the bar.
% In each of these previous works the prevention of \textit{progress reversal} is a challenge which remains out of scope, and this is one of the key problems we address in our work.

\section{Approach}
\label{sec:approach}

We investigate multi-step tasks for which there is a sparse and approximate notion of task progress.
It is possible to improve the efficiency of learning by taking these four measures: structuring such problems to capture invariant properties of the data, deploying traditional algorithms where they are most effective, ensuring rewards do not propagate through failed actions, and introducing an algorithm which removes unnecessary exploration.
We will later demonstrate our approach in the context of the general problem of assembly through vision-based robotic manipulation.

We frame the problem as a Markov Decision Process $(S, A, P, R)$, with state space $S$, action space $A$, transition probability function $P : S \times S \times A \rightarrow \mathbb{R}$, and reward function $R : S \times A \rightarrow \mathbb{R}$.
This includes a simplifying assumption equating sensor observations and state. % This isn't needed since everyone does this, but okay.
At time step $t$, the agent observes state $s_t$ and chooses an action $a_t$ according to its policy $\pi : S \rightarrow A$.
The action results in a new state $s_{t+1}$ with probability $P(s_{t+1} | s_t, a_t)$.
As in VPG~\cite{zeng2018vpg}, we use Q-learning to produce a deterministic policy for choosing actions.
The function $Q : S \times A \rightarrow \mathbb{R}$ estimates the expected reward $R$ of an action from a given state, i.e.~the ``quality'' of an action. 
Our policy $\pi$ selects an action $a_t$ as follows:
\begin{equation}
    % a_t \leftarrow 
    \pi(s_t) = \argmax_{a \in A} Q(s_t, a)
    \label{eq:policy_pi}
\end{equation}
Thus, the goal of training is to learn a $Q$ that maximizes $R$ over time.
This is accomplished by iteratively minimizing $|Q(s_t, a_t) - y_t|$, where the target value $y_t$ is:
\begin{equation}
    \label{eq:q-learning}
    y_t = R(s_{t+1}, a_t) + \gamma Q(s_{t+1}, \pi(s_{t+1}))
\end{equation}

Q-learning is a fundamental algorithm in RL, but there are key limitations in its most general form for applications like robotics where the space and cost of actions and new trials is extremely large, and efficient exploration can be essential or even safety critical.
It is also highly dependent on $R$, whose definition can cause learning efficiency to vary by orders of magnitude, as we show in Sec. \ref{subsec:algorithm_ablation}, and so we begin with our approach to reward shaping.

\subsection{Reward Shaping}
\label{sec:reward_shaping}
Reward shaping is an effective technique for optimizing a reward $R$ to train policies~\cite{ng1999policy} and their neural networks efficiently. 
Here, we present several reward functions for later comparison (Sec. \ref{subsec:algorithm_ablation}), which build towards a general formulation for reward shaping conducive to efficient learning on a broad range of novel tasks, thus reducing the \emph{ad hoc} nature of successful reward schedules. 
% Reward shaping is an effective technique for training these neural networks, which in our case involves rewarding intermediate sub-tasks $\Phi$, as well as the overall task.
% We will define several reward functions which incorporate different properties of the state and are  ``instantaneous'' in the midst of a trial, as well as others which require a completed trial. 
% In our ablation study (Sec. \ref{subsec:algorithm_ablation}) we will compare how each reward contributes to overall learning.

\begin{figure}[bt!]

    \vspace{0.4cm}
    % \centering
    % \hfill
    % \includegraphics[width=\columnwidth]{grasp_efficiency.pdf}
    % \includegraphics[width=\columnwidth]{2019-09-12-18-21-37-push-grasp-16k-trial-reward_success_plot}
    % \includegraphics[width=\columnwidth]{2020-02-07-14-43-44_Sim-Push-and-Grasp-Trial-Reward-Common-Sense-Training-Sim-Push-and-Grasp-SPOT-Reward-Common-Sense-Training_success_plot}
    % \includegraphics[width=\columnwidth]{example_of_spot_reward.png}
    % \includegraphics[width=\columnwidth]{example_of_spot_reward_plot_v2_cropped_compressed.pdf}
    \includegraphics[width=\columnwidth]{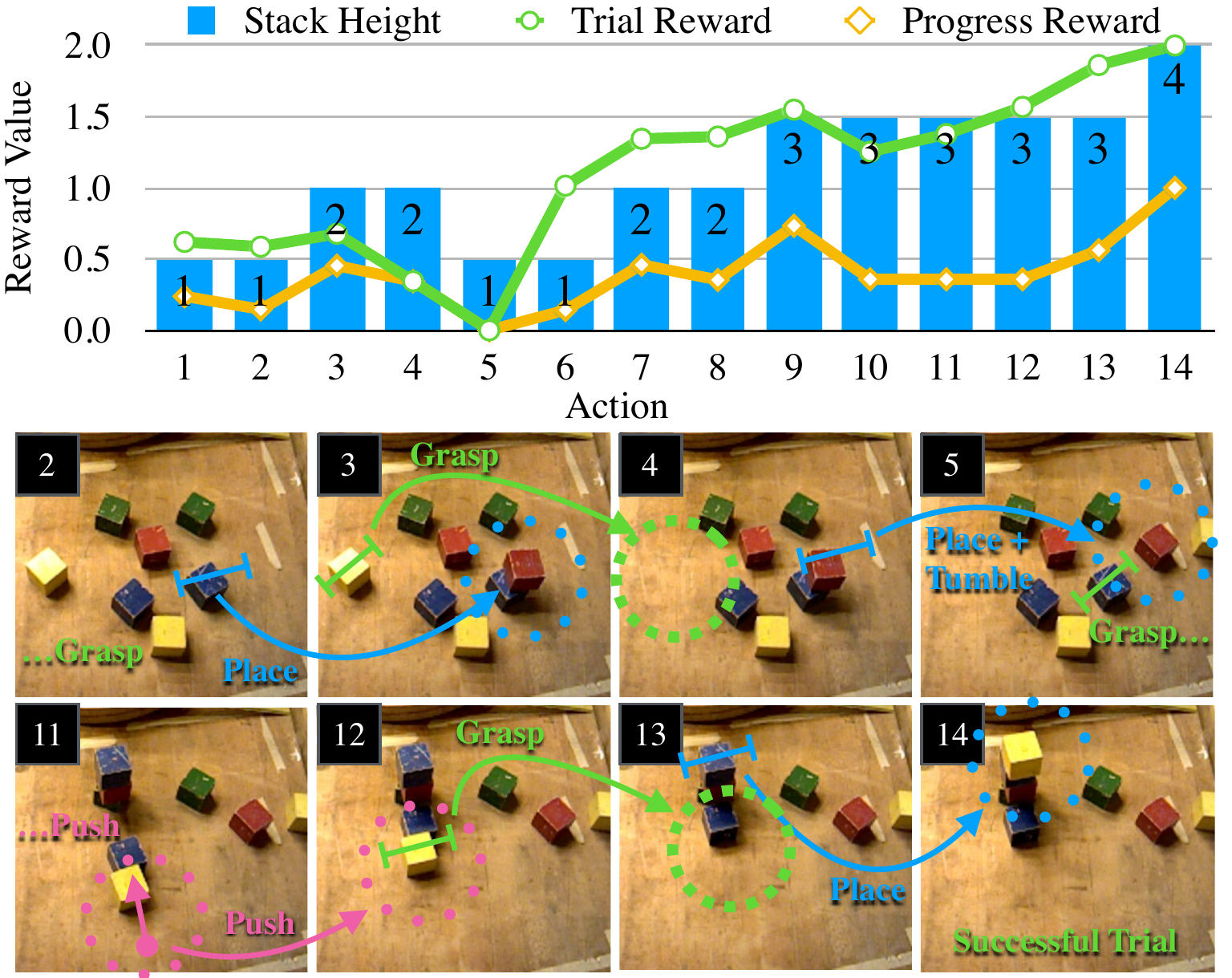}
    \caption{\label{fig:trial-reward}
Example of SPOT Trial Reward $\rtrial$ (eq. \ref{eq:trial_reward}), and the SPOT Progress Reward $\rprogress$ (eq. \ref{eq:instant_progress}) with images of key action steps. 
% With a reward of 0.5 for push, 1 for grasp, 2 for place, and $W=0.65$. 
% The color images captured after some highlight actions are shown. 
Actions 1-3: $a_1$ is an initial grasp, followed by a successful place where a slightly off balance stack of height 2 is formed.
Actions 4-5: \textit{Progress reversal} occurs when a grasp then place knocks the stack over, so the reward values go to zero.
Action 7: While not pictured, the scene is similar to $a_3$ but with a better balanced top block. 
Intuitively, since $a_9$ doesn't topple like $a_5$ a better reward at $a_7$ would be appropriate, which is one advantage of $\rtrial$ over $\rprogress$, because $\rprogress(s_4, a_3) = \rprogress(s_8, a_7)$ and $\rtrial(s_4, a_3) < \rtrial(s_8, a_7)$ because $a_7$ leads directly to a successful stack.
Actions 11-14: Grasp and place actions lead to a full stack of 4 completing the trial. 
The final $\rtrial$ at $a_{14}$ is $2\times \rprogress$.
Here $W_{\phi_t} \in  \{W_{push}\!=\!.5, W_{grasp}\!=\!1, W_{place}\!=\!1.25\}$ for chart visibility. % TOODO(ahundt) double check W values
    }
    \vspace{-0.5cm}
\end{figure}

Suppose each action $a$ is associated with a sub-task $\phi \in \Phi$ and that we have an indicator function $\1_{a}[s_{t+1}, a_t]$ which equals 1 if an action $a_t$ succeeds at $\phi$ and 0 otherwise\footnote{Examples of action indicator sources include the grasp detector in our Robotiq 2F85 gripper, human supervision, or another detection algorithm.}.
As in VPG~\cite{zeng2018vpg}, our baseline rewards follow this principle and include a sub-task weighting function $W : \Phi \rightarrow \mathbb{R}$, according to their subjective difficulty and importance\footnote{In our experiments we assign simple values for each successful action type: $W_{\phi_t} \in  \{W_{push}\!=\!0.1, W_{grasp}\!=\!1, W_{place}\!=\!1\}$.}:
% $W(\phi_t)$
% which determines the appropriate reward for each successful action type:
\begin{equation}
    \label{eq:vpg}
    R_{\text{base}}(s_{t+1}, a_t) = W(\phi_t) \1_{a}[s_{t+1}, a_t].
\end{equation}
Next, we define a sparse and approximate task progress function $\progress : S \rightarrow \mathbb{R}\in [0,1]$, indicating proportional progress towards an overall goal, where $\progress(s_t)=1$ means the task is complete\footnote{In our block tasks $\progress$ is the height of the stack or length of the row vs the goal size, in table clearing either the number of objects or occupied pixels vs the total, and in navigation the remaining vs initial distance.
}.
As in our story of Spot the dog (Sec. \ref{sec:introduction}), a \textit{progress reversal} leads us to perform Situation Removal (SR) on the agent and physically reset the environment during training (Fig. \ref{fig:gripper_asymmetry_time_dependencies}).
We define an associated indicator $\1_{\text{SR}}[s_{t}, s_{t+1}]$, which equals 1 if $\progress(s_{t+1}) \geq \progress(s_{t})$ and 0 otherwise.
These lead to new reward functions:
\begin{equation}
% \vspace{-0.5cm}
    \label{eq:situation_removal}
    R_{\text{SR}}(s_{t+1}, a_t) = \1_{\text{SR}}[s_{t},s_{t+1}]R_{\text{base}}(s_{t+1}, a_t)
\end{equation}
\begin{equation}
\label{eq:instant_progress}
\rprogress(s_{t+1}, a_t) =
%   \frac{\progress_t}{\progress_{\max}}R_{\text{SR}}(s_{t+1}, a_t)
   \progress(s_{t+1}) R_{\text{SR}}(s_{t+1}, a_t)
\end{equation}
One advantage of $R_{\text{base}}$, $R_{\text{SR}}$, and $\rprogress$ is that each is available ``instantaneously'' in the midst of a trial after two state transitions. 
However, they do not consider the possibility that an early mistake might lead to failure many steps down the line (Fig. \ref{fig:gripper_asymmetry_time_dependencies}, \ref{fig:trial-reward}), and so we will develop a reward which propagates across whole trials.

\subsection{Situation Removal: SPOT Trial Reward}
\label{subsec:spot_trial_reward}
Is it possible for a reward function to account for actions which lead to failures at a later time step while still training more efficiently than a standard discounted reward $\rdiscount$ where $\rdiscount(s_{t+1}, a_t)\!=\!\gamma \rdiscount(s_{t+2}, a_{t+1})$?
Our approach is to block reward propagation through failed actions via the Situation Removal concept:
\begin{equation}
\label{eq:trial_reward}
\rtrial(s_{t+1}, a_t) = 
\begin{cases}
  0, & \text{if } R_*(s_{t+1},a_t) = 0\\
  2R_*(s_{t+1}, a_t), & \text{if } t = N\\
%   R^t + \gamma R^{t+1}, & \text{else if}\ R^t > 0\\
  R_*(s_{t+1}, a_t) + \gamma & \hspace{-1em} \rtrial (s_{t+2}, a_{t+1}), \\
  & \text{otherwise}
\end{cases}
\end{equation}
where  $R_*$ can be an arbitrary instant reward function such as $R_{\text{SR}}$ or $\rprogress$ from Sec. \ref{sec:reward_shaping}, $N$ marks the end of the trial, and $\gamma$ is the usual discount factor which is set to $\gamma=0.65$. 
% These values are recursively rolled out from the final action to the first action of a trial.
The effect of using $\rtrial$ is that future rewards only propagate across time steps where subtasks are completed successfully. 
As illustrated in Fig.~\ref{fig:trial-reward} and described in the caption,
the zero reward from situation removal cuts the propagation of future rewards back through time steps containing failed actions.
This focuses learning on short and successful sequences that complete a task.

% EMPTY FIGURE HACK TO SHIFT ALGORITHM 1 DOWN AND FIT IN MARGINS!
\begin{figure}
\vspace{0.1cm}
\end{figure}
\begin{algorithm}[bt]
\caption{SPOT-Q with Prioritized Experience Replay}
\label{alg:spot_q_replay}
\begin{algorithmic}[1]
\State \textbf{Input} Replay Memory $H_T\!=\!(S_T,A_T,R_T$,$\mathrm{Predicted}_T)$
\While {$\Call{agent\_is\_acting}{ }$}
    \State $t = \Call{prioritized\_experience\_sample}{T, H_T}$
    \State $y_t = R(s_{t+1}, a_{t}) + \gamma Q(s_{t+1}, \pi(s_{t+1}))$
    \State $\delta_t = \Call{huber\_loss}{Q(s_t,a_t) ;\;y_t}$
    \State $a_{\pi, t} = \pi(s_t)$
    \If{$M(s_t, a_{\pi,t})=0$} \Comment The action would fail.
        \State $y'_t = \gamma Q(s_{t+1},a_{\pi,t} )$ \Comment New 0 reward sample.
        \State $\delta_t = \delta_t + \Call{huber\_loss}{Q(s_t,a_t) ;\;y'_t}$
    \EndIf
    \State $\Call{backprop}{\sum{}{\delta_t}}$; step optimizer; update weights.
\EndWhile
\end{algorithmic}
\end{algorithm}

\subsection{SPOT-Q Learning and Dynamic Action Spaces}
\label{sec:common_sense}

%Due to high hardware costs in robotics it is desirable to utilize time as efficiently as possible.
In this section, we go a step further and leverage \emph{a priori} knowledge about the environment to make simple but powerful assumptions which both reduce unproductive attempts and accelerate training.
% TODO(ahundt) change predictive Mask to predictive oracle
Specifically, there are many occasions when certain action failures are easily predicted from the same sensor signal used for Q learning.  
To this end, we assume the existence of an oracle,
$M(s_t, a) \rightarrow \{0,1\},$ which takes the \emph{current} state $s_t$ and an action $a$ and returns 0 if an action is certain to fail, and 1 otherwise. 
%More generally, t could be a hand-tuned rule, as we show in Sec.~\ref{sec:experiments}, or the output of an instance segmentation model, \textit{e.g.}~\cite{xie2019best}.
%
This is subtly different from the success indicator $\1_{a}[s_{t+1}, a_t]$, which requires the \emph{outcome} $s_{t+1}$ of an action $a_t$ to determine success or failure\footnote{For example, grasping an object can only succeed if there is depth data in the neighborhood of a predicted action, so attempts to grasp in free space can be easily predicted to fail, as we demonstrate in Sec.~\ref{sec:experiments}.}.
Using $M$, we define the dynamic action space $M_t(A)$:
\begin{equation}
    \label{eq:dynamic-action-space}
    M_t(A) = \{a \in A | M(s_t, a) = 1\}.
\end{equation}
In short, $M_t(A)$ does not tell us whether $a\in A$ is an action worth taking, but rather whether it is worth exploring.

Given a state $s_t$, the question becomes how to most effectively utilize $M_t$ in training. If $\pi(s_t) \not\in M_t(A)$, then $\pi(s_t)$ can be treated as a failure for the purposes of learning \emph{and} we can explore the next best action not guaranteed to fail. To formalize this, we introduce \textbf{SPOT-Q Learning} which is a new target value function replacing eq. \ref{eq:q-learning}:
\begin{equation}
\label{eq:spotq}
y_{M, t} = 
\begin{cases}
y_t, & \text{if } \pi(s_{t+1}) \in M_t(A) \\
y_t + \gamma Q(s_{t+1}, \pi_M(s_{t+1})) & \text{otherwise}\\
\hspace{1.1em} + R(s_{t+1}, a_t), \\
\end{cases}
\end{equation}
where $\pi_M (s_t) = \argmax_{a \in M_t(A)} Q(s_t, a)$.
Crucially, we perform backpropagation on both the masked action, which has 0 reward, and the unmasked action $\pi_M(s_t)$, which the robot actually performs.

% That means when $\argmax_{a \in A}(Q(s_t,a)) \not\in M_t(A)$ we get a second bonus coordinate where we can perform backpropagation on two actions simultaneously.
% That means when the chosen action is masked we get a second bonus action and we perform backpropagation on two actions simultaneously.
Alg. \ref{alg:spot_q_replay} describes how we continuously train from past examples with SPOT-Q and Prioritized Experienced Replay (PER)~\cite{schaul2016prioritized} as the current policy is rolled out.
% Even as $\pi$ changes, SPOT-Q is re-applied for each new choice of action, so we continuously get new zero reward labels from every replay sample where $\pi(s_t) \not\in M_t(A)$.
In Sec.~\ref{sec:experiments}, we discuss how SPOT-Q allows us to surpass prior work, wherein similar heuristics~\cite{zeng2018vpg,2018qtopt} neither match SPOT-Q nor account for the safety considerations we discuss later.
% covered in Sec. \ref{sec:lava_gridworld_safety_domain_generalization}.

\section{Simulation Experiments}
\label{sec:experiments}

%In this section, we show that o
Our method improves performance and action efficiency over the state of the art on the table clearing task from VPG~\cite{zeng2018vpg}, as well as on two challenging multi-step tasks of our design: creating a stack of four blocks and creating a horizontal row of four blocks. Our best results can achieve 100\% trial success on the simulated stacking and row tasks, models which successfully transfer to the real world as we show in Sec. \ref{subsec:real_robot_experiments}.

We detail a series of simulation experiments to understand the contribution of each element of our approach to this overall performance. %the algorithms introduced in Sec. \ref{sec:approach}.
%We then move more realistic simulated robots in CoppeliaSim aka V-REP where we first evaluate the baseline pushing and grasping table clearing scenarios provided by VPG~\cite{zeng2018vpg}.
%Most important, we evaluate two multi-step tasks of our own design: 4 cube stacking and row-making.
To do so, we evaluate each reward function, the effect of SPOT-Q on heuristic exploration, other possible SPOT-Q implementations, the reward weighting term $W$, and then we describe our best results with SPOT-Q + $\rprogress$ and SPOT-Q + $\rtrial$. 
In brief, we find that Situation Removal $R_{\text{SR}}$ is the largest contributor to our improved performance, $\rprogress$ improves accuracy and efficiency, and $\rtrial$ trains more efficiently than discounted rewards while accounting for a time delay between actions and consequences.
%SPOT-Q (eq. \ref{eq:spotq}) improves results over no masking, and naive masking (eq. \ref{eq:dynamic-action-space}).
SPOT-Q improves results over no masking, and over basic masking on its own.
Finally, we test a grid world navigation task~\cite{gym_minigrid} 
%to show convergence including a preliminary evaluation of how SPOT-Q applies to safe exploration, to show the broad applicability of these algorithms.
to show how the SPOT framework applies to safe reinforcement learning.
Tables \ref{table:StackingResults} and \ref{table:gridworld_results} summarize these results. %, with descriptions and analysis below.

\subsection{Robot Implementation Details}
\label{subsec:robot_implementation_details}
We consider a robot capable of
being commanded to a specified arm pose and gripper state in its workspace.
Our action space consists of three components: action types $\Phi$, locations $X \times Y$, and angles $\Theta$.
The agent observes the environment via a fixed RGB-D camera, which we project so that $z$ is aligned with the direction of gravity, as shown in Fig.~\ref{fig:stack_EVT_q_network}.
We discretize the spatial action space into a square height map with $0.448 m$ on a side and $224\times 224$ bins with coordinates $(x, y)$, so each pixel represents roughly $4 mm^2$ as per VPG~\cite{zeng2018vpg}.
The angle space $\Theta = \{\frac{2\pi i}{k} | i \in [0, k - 1]\}$ is similarly discretized into $k=16$ bins.
The set of action types consists of three high-level motion primitives $\Phi = \{\mathtt{grasp}, \mathtt{push}, \mathtt{place}\}$.
In our experiments action success is determined by our gripper's sensor for $\mathtt{grasp}$, object perturbations for $\mathtt{push}$, and an increase in stack height or row length for $\mathtt{place}$.

A traditional trajectory planner executes each action $a = (\phi, x, y, \theta) \in A$ on the robot. 
% The gripper $z$ position is a fixed offset relative to the heightmap.
For grasping and placing, each action moves to $(x,y)$ with gripper angle $\theta \in \Theta$ and closes or opens the gripper, respectively. 
A push starts with the gripper closed at $(x,y)$ and moves horizontally a fixed distance along angle $\theta$.
Fig.~\ref{fig:stack_EVT_q_network} visualizes our overall algorithm, including the action space and corresponding $Q$-values.

\subsection{Evaluation Metrics}
\label{sec:metrics}
We evaluate our algorithms in randomized test cases in accordance with the metrics found in VPG~\cite{zeng2018vpg}.
% These include the percentage of successful grasps, placement action efficiency, and trial completion rate.
% The completion rate is defined as the percentage of trials where the policy is able to successfully complete a task before the grasp or push action fails 10 consecutive times.
% Success of a push is when more than 300 depth pixels have changed in a scene. 
% A successful grasp is counted upon two consecutive detections by the internal Robotiq grasp sensor, once upon closing and once after lifting.
% A successful place for stacking is evaluated more specifically by the maximum $z$ height in the heightmap. 
% Alternately it can be awarded when height the highest vertical $z$ height of a scene has increased by a minimum threshold.
Ideal Action Efficiency is 100\% and calculated as $\mathtt{Ideal} / \mathtt{Actual}$ action count; defined as 1 action per object for grasping tasks; and 2 actions per object for tasks which involve placement.
This means 6 total actions for a stack of height 4 since only 3 objects must move, and 4 total actions for rows by placing two blocks between two endpoints.
We validate simulated results twice with 100 trials of novel random object positions.

\subsection{Algorithm Ablation}
\label{subsec:algorithm_ablation}
We compare the contribution from each component of the underlying algorithm and against baseline approaches in Table \ref{table:StackingResults}, except for clearing tasks which are provided in the text.
Unless otherwise stated we summarize rows and stacks together as a combined average below.

\textbf{Clear 20 Toys:} We establish a baseline via the primary simulated experiment found in VPG~\cite{zeng2018vpg}, where 20 toys with varied shapes must be grasped to clear the robot workspace.
% Some differences occur because we also account for our asymmetric gripper, which means training cannot automatically be applied at the both the actual gripper rotation angle and its 180\degree offset. Fig.~\ref{fig:gripper_asymmetry_time_dependencies} shows such considerations in detail.
% Additionally, our training performs experience replay in parallel with robot actions. 
The SPOT framework matches VPG~\cite{zeng2018vpg} with 100\% task completion and improves both the rate of grasp successes from 68\% to 84\% and action efficiency from 64\% to 74\%.
% EVT reduces the Action Efficiency Error (1 - Action Efficiency) from 39\% with VPG to 14\% with EVT. % formerly 18.3\%

\textbf{Clear Toys Adversarial:} The second baseline scenario is the 11 cases of challenging adversarial arrangements from VPG~\cite{zeng2018vpg}, where toys are placed in tightly packed configurations. 
Each case is run 10 times and the SPOT framework completely clears 7/11 cases compared to 5/11 in VPG~\cite{zeng2018vpg}; the clearance rate across all 110 runs improves to 95\% from 84\%. 
Efficiency in this case drops from 60\% to 38\%, which is accounted for by the increase in the number of difficult cases solved, as separating the blocks can take several attempts.

\textbf{Reward Functions:}  $R_{\text{base}}$, $R_{\text{SR}}$, $\rprogress$, and $\rtrial$ incrementally extend one another (Sec. \ref{sec:reward_shaping}, \ref{subsec:spot_trial_reward}). 
All masking is disabled for this study unless otherwise indicated.

$\rdiscount$ s.t. $\rdiscount(s_{t+1}, a_t) = \gamma \rdiscount(s_{t+2}, a_{t+1})$ 
is discounting, the most conventional approach to trial rewards. 
When evaluated with $\rprogress$ at the final time step and $\gamma=0.9$,
% , since ideal stacking trials should only take 6 actions.
grasp and place actions succeed at a rate of 5\% and 45\%, respectively. 
Stacks of height 2-3 are created and performance improves with masking (32\%, 48\%).
However, this approach is incredibly inefficient with no stacks of 4 within 20k actions. 
That said, we would expect convergence if orders of magnitude more training were feasible~\cite{andrychowicz2017hindsight}.

$R_{\text{base}}$ is effective for pushing and grasping~\cite{zeng2018vpg}, but it is not sufficient for multi-step tasks, only completing 13\% of rows and stacks with about $200+$ actions per trial in the best case. 
In another case, it repeatedly reverses progress by often looping grasping then placing of the same object at one spot, leading to 99\% successful grasps but 0 successful trials overall, even after manual scene resets.
We do not expect $R_{\text{base}}$ to converge on these tasks as there is no progress signal to indicate, for example, that grasping from the top of an existing stack is a poor choice.

% RSR trial success (90+95+94+98)/4 =94.25     action efficiency (8+23+19+43)/4 =23.25
$R_{\text{SR}}$ resolves the progress reversal problem immediately since such actions get 0 reward; and thus we see an astounding increase in trial successes from 13\% to 94\%, and an order of magnitude efficiency increase to 23\% across both tasks, or about $22$ actions per trial.

% Rprogress trial success (98+98+96+98)/4=97     efficiency (38+52+34+57)/4=45
$\rprogress$ leads to a rise in combined trial successes to 97\%, and efficiency to 45\%, or about 20 actions per trial. 
This improves upon pure situation removal by incorporating the quantitative amount of progress.

$\rtrial$ utilizes $\rprogress$ as the instant reward function in this test, and has an average trial success rate of 96\% for stacks and efficiency of 31\%, or about $19$ actions per trial.
However, performance degrades significantly for rows, declining to an 80\% trial success rate and just 16\% action efficiency, or about 25 actions per trial.
These values indicate $\rtrial$ strikes a trade-off between the inefficiency of $\rdiscount$ and the need for a more instantaneous progress metric in $\rprogress$, as the most recent value can be utilized to fill actions with no progress feedback.
We also note that once SPOT-Q is added this reward is the best for stacking and second best overall, as we show below.

\begin{table}\centering
\vspace{0.4cm}
\setlength\tabcolsep{1.5pt} % default value: 6pt
% \ra{1.3}
\begin{tabular}{LCCCCCCCCRRRR@{}}\toprule
% \rowcolor{white}\textbf{Simulation} & Algorithm &Action& Trials & Action\\
% \rowcolor{white}Test Task& Ablation &Space& Complete & Efficiency\\
\rowcolor{white}\textbf{Simulation Stack of 4} &SPOT-Q&Mask&Reward& Trials & Efficiency\\
\midrule
 Discounted $R_t\!=\!\gamma R_{t+1}$ &\xmark&\xmark&$\rdiscount$& 0\%&0\%\\ % femur '/home/ahundt/src/real_good_robot/logs/2020-05-12-16-42-47_Sim-Stack-Discounted-Reward-Training'
\rowcolor{white} Discounted $R_t\!=\!\gamma R_{t+1}$  &\xmark&\cmark&$\rdiscount$& 0\%&0\%\\ % femur  '/home/ahundt/src/real_good_robot/logs/2020-05-12-14-41-34_Sim-Stack-Two-Step-Reward-Common-Sense-Training/best_stats.json'
% /home/ahundt/src/real_good_robot/logs/2020-05-22-20-49-56_Sim-Stack-Two-Step-Reward-Training/ only 2 trials completed successfully during 20k actions of training with a max training efficiency of 1%. This problem is repeatable.
% baseline stack costar logs/2020-04-25-21-41-04_Sim-Stack-Two-Step-Reward-Training/2020-04-27-17-44-27_Sim-Stack-Two-Step-Reward-Testing/best_stats.json {"action_efficiency_best_index": 3991, "action_efficiency_best_value": 0.019543973941368076, "grasp_success_rate_best_index": 3991, "grasp_success_rate_best_value": 0.9404958677685951, "place_success_rate_best_index": 3991, "place_success_rate_best_value": 0.5837526959022286, "trial_success_rate_best_index": 3991, "trial_success_rate_best_value": 0.13}
Baseline $R_{\text{base}}$ eq. \ref{eq:vpg} &\xmark&\xmark&$R_{\text{base}}$ & 2-13\% & 1-2\%\\ % costar 2020-04-25-21-41-04_Sim-Stack-Two-Step-Reward-Training/2020-04-27-17-44-27_Sim-Stack-Two-Step-Reward-Testing/best_stats.json
% SITUATION REMOVAL femur 2020-05-22-14-57-54_Sim-Stack-Two-Step-Reward-Training/2020-05-28-12-51-52_Sim-Stack-Two-Step-Reward-Testing femur {'grasp_success_rate_best_value': 0.8149812734082397, 'grasp_success_rate_best_index': 2454, 'place_success_rate_best_value': 0.6660714285714285, 'trial_success_rate_best_index': 2454, 'action_efficiency_best_index': 2456, 'place_success_rate_best_index': 2454, 'trial_success_rate_best_value': 0.95, 'action_efficiency_best_value': 0.23471882640586797}
% '/home/ahundt/src/real_good_robot/logs/2020-05-30-13-11-57_Sim-Stack-Two-Step-Reward-Training/2020-06-02-19-24-53_Sim-Stack-Two-Step-Reward-Testing' {"action_efficiency_best_index": 7161, "action_efficiency_best_value": 0.07710574102528286, "grasp_success_rate_best_index": 7159, "grasp_success_rate_best_value": 0.8780807551127425, "place_success_rate_best_index": 7159, "place_success_rate_best_value": 0.6434548714883442, "trial_success_rate_best_index": 7159, "trial_success_rate_best_value": 0.9} spot workstation
\rowcolor{white}Situation Rem. $R_{\text{SR}}$ eq. \ref{eq:situation_removal} &\xmark&\xmark&$R_{\text{SR}}$& 90-95\% & 8-23\%\\ % 

Task Prog. $\rprogress$ eq. \ref{eq:instant_progress} &\xmark&\xmark &$\rprogress$& 98-98\% & 38-52\%\\

\rowcolor{white}Trial Rew. $\rtrial$ eq. \ref{eq:trial_reward} &\xmark&\xmark &$\rtrial$& 95-97\% & 30-32\%\\
% rtrial mask no-spotq femur 2020-05-23-14-31-09_Sim-Stack-SPOT-Trial-Reward-Masked-Training/2020-05-27-04-58-39_Sim-Stack-SPOT-Trial-Reward-Masked-Testing femur trial success rate 0.99 action efficiency 0.55    > Random Testing results:    >  {'trial_success_rate_best_value': 0.99, 'trial_success_rate_best_index': 1082, 'grasp_success_rate_best_value': 0.8155668358714044, 'grasp_success_rate_best_index': 1082, 'place_success_rate_best_value': 0.790650406504065, 'place_success_rate_best_index': 1082, 'action_efficiency_best_value': 0.5545286506469501, 'action_efficiency_best_index': 1084}
% rtrial mask no-spotq > Random Testing Complete! Dir: /home/pittsburgh/src/real_good_robot/logs/2020-06-01-16-31-02_Sim-Stack-SPOT-Trial-Reward-Masked-Training/2020-06-07-17-25-28_Sim-Stack-SPOT-Trial-Reward-Masked-Testing > Random Testing results: >  {'trial_success_rate_best_value': 0.95, 'trial_success_rate_best_index': 1242, 'grasp_success_rate_best_value': 0.8168389955686853, 'grasp_success_rate_best_index': 1242, 'place_success_rate_best_value': 0.7597173144876325, 'place_success_rate_best_index': 1242, 'action_efficiency_best_value': 0.463768115942029, 'action_efficiency_best_index': 1244}
Mask but no SPOT-Q &\xmark&\cmark&$\rtrial$& 95\%-99\% & 46\%-55\% \\
\rowcolor{white}\textbf{SPOT-Q + $\rtrial$}&\cmark&\cmark&\boldmath$\rtrial$& \textbf{100-100\%} & \textbf{45-51\%}\\
% stack rprogress spot-q spot workstation Random Testing Complete! Dir: /home/ahundt/src/real_good_robot/logs/2020-06-03-11-44-02_Sim-Stack-Two-Step-Reward-Masked-Training/2020-06-07-06-26-25_Sim-Stack-Two-Step-Reward-Masked-Testing Random Testing results: {'trial_success_rate_best_value': 1.0, 'trial_success_rate_best_index': 1351, 'grasp_success_rate_best_value': 0.7558746736292428, 'grasp_success_rate_best_index': 1351, 'place_success_rate_best_value': 0.757679180887372, 'place_success_rate_best_index': 1351, 'action_efficiency_best_value': 0.44855662472242785, 'action_efficiency_best_index': 1353}
% stack rprogress spot-q spot workstation random testing     > Random Testing Complete! Dir: /home/ahundt/src/real_good_robot/logs/2020-06-04-11-18-49_Sim-Stack-Two-Step-Reward-Masked-Training/2020-06-08-05-43-46_Sim-Stack-Two-Step-Reward-Masked-Testing    > Random Testing results:    > {'trial_success_rate_best_value': 0.96, 'trial_success_rate_best_index': 2319, 'grasp_success_rate_best_value': 0.732488822652757, 'grasp_success_rate_best_index': 2320, 'place_success_rate_best_value': 0.693564862104188, 'place_success_rate_best_index': 2321, 'action_efficiency_best_value': 0.25097024579560157, 'action_efficiency_best_index': 2321}

\textbf{SPOT-Q + }$\rprogress$&\cmark&\cmark&\boldmath$\rprogress$& \textbf{96-100\%} & \textbf{25-45\%}\\
% 2020-02-16-22-38-38_Sim-Stack-SPOT-Trial-Reward-Testing DenseNet, on CoSTAR, loose action boundaries
% \rowcolor{white}\textbf{Real}&Stack of 4 Cubes& SPOT (eq. \ref{eq:trial_reward}) &\cmark (eq. \ref{eq:spotq})& \textbf{82\%} & \textbf{71\%} & \textbf{82\%} & \textbf{60\%}\\ % 2020-02-09-11-02-57_Real-Stack-SPOT-Trial-Reward-Common-Sense-Training
\midrule
\rowcolor{white}\textbf{Simulation Row of 4} &SPOT-Q&Mask&Reward& Trials & Efficiency\\
\midrule
% baseline costar 2020-04-25-21-41-35_Sim-Rows-Two-Step-Reward-Training/2020-04-27-17-25-31_Sim-Rows-Two-Step-Reward-Testing/best_stats.json 13% trial 1% efficiency
% femur /home/ahundt/src/real_good_robot/logs/2020-05-25-14-13-02_Sim-Rows-Two-Step-Reward-Training 1 training trial success ~0\% eff ~0%\%,  first trial took 1200 actions, manually resetting to the next trial didn't improve, so rounding down to 0.
Baseline $R_{\text{base}}$ eq. \ref{eq:vpg} &\xmark&\xmark&$R_{\text{base}}$& 0-13\% & 0-1\%\\
% femur '/home/ahundt/src/real_good_robot/logs/2020-05-18-20-27-01_Sim-Rows-Two-Step-Reward-Training/2020-05-24-16-22-22_Sim-Rows-Two-Step-Reward-Testing/best_stats.json' Situation Removal 89% trial success 15% action efficiency
% spot workstation rsr basic situation removal no-mask  '/home/ahundt/src/real_good_robot/logs/2020-05-30-17-46-01_Sim-Rows-Two-Step-Reward-Training/2020-06-03-06-28-31_Sim-Rows-Two-Step-Reward-Testing' {"action_efficiency_best_index": 2092, "action_efficiency_best_value": 0.28995215311004785, "grasp_success_rate_best_index": 2090, "grasp_success_rate_best_value": 0.8680926916221033, "place_success_rate_best_index": 2090, "place_success_rate_best_value": 0.5927835051546392, "trial_success_rate_best_index": 2090, "trial_success_rate_best_value": 0.94}
% > Random Testing Complete! Dir: /home/ahundt/src/real_good_robot/logs/2020-06-07-14-06-58_Sim-Rows-Two-Step-Reward-Training/2020-06-10-23-16-04_Sim-Rows-Two-Step-Reward-Testing    > Random Testing results:     >  {'trial_success_rate_best_value': 0.98, 'trial_success_rate_best_index': 1456, 'grasp_success_rate_best_value': 0.8350125944584383, 'grasp_success_rate_best_index': 1457, 'place_success_rate_best_value': 0.60422, 'place_success_rate_best_index': None, 'action_efficiency_best_value': 0.4368131868131868, 'action_efficiency_best_index': 1458}
% Rows Efficiency conversion 6 actions ideal -> 4 actions ideal 28.9*4/6=19.2; 43.6*4/6=29.0
\rowcolor{white}Situation Rem. $R_{\text{SR}}$ eq. \ref{eq:situation_removal} &\xmark&\xmark&$R_{\text{SR}}$& 94-98\% & 19-43\%\\ % 
% Rows Trial Data: https://github.com/jhu-lcsr/costar_visual_stacking/releases/tag/v0.12.0
% rprogress workstation named spot '/home/ahundt/src/real_good_robot/logs/2020-05-30-13-10-52_Sim-Rows-Two-Step-Reward-Training/2020-06-03-01-45-49_Sim-Rows-Two-Step-Reward-Testing/best_stats.json' {"action_efficiency_best_index": 1186, "action_efficiency_best_value": 0.5118243243243243, "grasp_success_rate_best_index": 1184, "grasp_success_rate_best_value": 0.9146537842190016, "place_success_rate_best_index": 1185, "place_success_rate_best_value": 0.8300884955752212, "trial_success_rate_best_index": 1184, "trial_success_rate_best_value": 0.96}
% costar > Random Testing Complete! Dir: /media/costar/f5f1f858-3666-4832-beea-b743127f1030/real_good_robot/logs/2020-06-01-13-03-15_Sim-Rows-Two-Step-Reward-Training/2020-06-05-09-17-42_Sim-Rows-Two-Step-Reward-Testing Random Testing results: {'trial_success_rate_best_value': 0.98, 'trial_success_rate_best_index': 700, 'grasp_success_rate_best_value': 0.8403141361256544, 'grasp_success_rate_best_index': 700, 'place_success_rate_best_value': 0.8463949843260188, 'place_success_rate_best_index': 700, 'action_efficiency_best_value': 0.8657142857142858, 'action_efficiency_best_index': 702}
% Rows Efficiency conversion 6 actions ideal -> 4 actions ideal 51.1*4/6=34.0; 86.6*4/6=57.7
Task Prog. $\rprogress$ eq. \ref{eq:instant_progress} &\xmark&\xmark &$\rprogress$&  96-98\% & 34-57\%\\ 
% > Random Testing Complete! Dir: /home/costar/src/real_good_robot/logs/2020-05-18-19-57-17_Sim-Rows-SPOT-Trial-Reward-Training/2020-05-23-16-29-39_Sim-Rows-SPOT-Trial-Reward-Testing    > Random Testing results:    > {'trial_success_rate_best_value': 0.87, 'trial_success_rate_best_index': 3413, 'grasp_success_rate_best_value': 0.8074894514767933, 'grasp_success_rate_best_index': 3413, 'place_success_rate_best_value': 0.6021080368906456, 'place_success_rate_best_index': 3413, 'action_efficiency_best_value': 0.1652505127453853, 'action_efficiency_best_index': 3413}
%  > '/home/ahundt/src/real_good_robot/logs/2020-05-28-12-10-18_Sim-Rows-SPOT-Trial-Reward-Training/2020-06-11-00-40-54_Sim-Rows-SPOT-Trial-Reward-Testing/best_stats.json'    > {"action_efficiency_best_index": 1779, "action_efficiency_best_value": 0.29375351716375914, "grasp_success_rate_best_index": 1778, "grasp_success_rate_best_value": 0.33109619686800895, "place_success_rate_best_index": null, "place_success_rate_best_value": 0.874, "trial_success_rate_best_index": 1777, "trial_success_rate_best_value": 0.74}
% Rows Efficiency conversion 6 actions ideal -> 4 actions ideal 16.5*4/6=11.0; 29.4*4/6=19.6
\rowcolor{white}Trial Rew. $\rtrial$ eq. \ref{eq:trial_reward} &\xmark&\xmark&$\rtrial$& 74-87\% & 11-20\%\\
% /home/costar/src/real_good_robot/logs/2020-05-24-09-36-39_Sim-Rows-SPOT-Trial-Reward-Masked-Training/2020-05-28-03-27-32_Sim-Rows-SPOT-Trial-Reward-Masked-Testing/best_stats.json {"action_efficiency_best_index": 1189, "action_efficiency_best_value": 0.4894869638351556, "grasp_success_rate_best_index": 1189, "grasp_success_rate_best_value": 0.7434402332361516, "place_success_rate_best_index": 1189, "place_success_rate_best_value": 0.8452380952380952, "trial_success_rate_best_index": 1189, "trial_success_rate_best_value": 0.93}
% > Random Testing Complete! Dir: /home/pittsburgh/src/real_good_robot/logs/2020-06-01-16-31-10_Sim-Rows-SPOT-Trial-Reward-Masked-Training/2020-06-06-20-50-48_Sim-Rows-SPOT-Trial-Reward-Masked-Testing > Random Testing results:  >  {'trial_success_rate_best_value': 0.92, 'trial_success_rate_best_index': 2371, 'grasp_success_rate_best_value': 0.5127226463104325, 'grasp_success_rate_best_index': 2372, 'place_success_rate_best_value': 0.7752808988764045, 'place_success_rate_best_index': 2373, 'action_efficiency_best_value': 0.24293547026571066, 'action_efficiency_best_index': 2373}
% Rows Efficiency conversion 6 actions ideal -> 4 actions ideal 48.9*4/6=32.6; 24.4*4/6=16.3
Mask but no SPOT-Q &\xmark&\cmark&$\rtrial$& 92-93\% & 16-32\% \\
% THIS LINE NOTE IS THE EFFICIENTNET MODEL FROM 2019-09, action efficiency is 44% assuming ideal 6 actions per row, 29% for 4 actions per row
% row spot-q rtrial costar > Random Testing Complete! Dir: /media/costar/f5f1f858-3666-4832-beea-b743127f1030/real_good_robot/logs/2020-05-13-12-21-00_Sim-Rows-SPOT-Trial-Reward-Masked-Training/2020-05-17-13-08-59_Sim-Rows-SPOT-Trial-Reward-Masked-Testing > Random Testing results: {"action_efficiency_best_index": 1789, "action_efficiency_best_value": 0.3764705882352941, "grasp_success_rate_best_index": 1785, "grasp_success_rate_best_value": 0.6561922365988909, "place_success_rate_best_index": 1785, "place_success_rate_best_value": 0.7634560906515581, "trial_success_rate_best_index": 1787, "trial_success_rate_best_value": 0.94}
% row spot-q rtrial femur Random Testing Complete! Dir: /home/ahundt/src/real_good_robot/logs/2020-05-06-20-58-40_Sim-Rows-SPOT-Trial-Reward-Common-Sense-Training/2020-05-09-22-03-51_Sim-Rows-SPOT-Trial-Reward-Common-Sense-Testing Random Testing results: {'trial_success_rate_best_index': 1216, 'grasp_success_rate_best_value': 0.6693989071038251, 'place_success_rate_best_value': 0.7818930041152263, 'action_efficiency_best_index': 1218, 'place_success_rate_best_index': 1216, 'trial_success_rate_best_value': 0.94, 'grasp_success_rate_best_index': 1216, 'action_efficiency_best_value': 0.5180921052631579}
% Rows Efficiency conversion 6 actions ideal -> 4 actions ideal 37.6*4/6=25.1; 51.8*4/6=34.5
\rowcolor{white}\textbf{SPOT-Q + $\rtrial$}&\cmark&\cmark&\boldmath$\rtrial$& \textbf{94-94\%} & \textbf{25-34\%}\\
% row spot-q rprogress on spot workstation Random Testing Complete! Dir: /home/ahundt/src/real_good_robot/logs/2020-06-03-12-05-28_Sim-Rows-Two-Step-Reward-Masked-Training/2020-06-06-21-34-07_Sim-Rows-Two-Step-Reward-Masked-Testing Random Testing results: {'trial_success_rate_best_value': 1.0, 'trial_success_rate_best_index': 667, 'grasp_success_rate_best_value': 0.850415512465374, 'grasp_success_rate_best_index': 667, 'place_success_rate_best_value': 0.7752442996742671, 'place_success_rate_best_index': 667, 'action_efficiency_best_value': 0.9265367316341829, 'action_efficiency_best_index': 667}
% row spot-q rprogress on spot workstation> Random Testing Complete! Dir: /home/ahundt/src/real_good_robot/logs/2020-06-03-23-18-31_Sim-Rows-Two-Step-Reward-Masked-Training/2020-06-07-17-17-16_Sim-Rows-Two-Step-Reward-Masked-Testing    > Random Testing results:    >  {'trial_success_rate_best_value': 0.98, 'trial_success_rate_best_index': 614, 'grasp_success_rate_best_value': 0.9190031152647975, 'grasp_success_rate_best_index': 614, 'place_success_rate_best_value': 0.7627118644067796, 'place_success_rate_best_index': 615, 'action_efficiency_best_value': 1.01628664495114, 'action_efficiency_best_index': 616}
% Rows Efficiency conversion 6 actions ideal -> 4 actions ideal 92.7*4/6=61.8; 101.6*4/6=67.7
\textbf{SPOT-Q + }$\rprogress$&\cmark&\cmark&\boldmath$\rprogress$& \textbf{98-100\%} & \textbf{62-68\%}\\
% NOTE THIS IS THE DENSENET MODEL FROM 2020-02-10-18-38-48_Sim-Rows-SPOT-Trial-Reward-Common-Sense-Training action efficiency is 86% assuming ideal 6 actions per row, 57% assuming ideal 4 actions per row.

% \midrule
% \rowcolor{white}Stack Toys 0.2m &  SPOT (eq. \ref{eq:trial_reward}) &\cmark (eq. \ref{eq:spotq})& 24\%& 55\% & 48\% & 4\%\\ % 2020-02-10-19-09-09_Sim-Stack-SPOT-Trial-Reward-Common-Sense-Training
% \midrule
\bottomrule
\end{tabular}
\caption{\label{table:StackingResults} Multi-step task test success rates measured out of 100\% for simulated tasks involving push, grasp and place actions trained for 20k actions (Sec. \ref{subsec:algorithm_ablation}).
Bold entries highlight our key algorithm improvements over the baseline.
``Trials'' indicates the overall rate at which stacks or rows are successfully completed.
``Efficiency'' is the test \texttt{Ideal/Actual} actions per trial.
The algorithm components are described in Sec. \ref{sec:approach}, except for ``Mask but no SPOT-Q'' which is a special case described in the SPOT-Q section of our ablation study (Sec. \ref{subsec:algorithm_ablation}).
Values are the min and max of two runs.
% Definitions: D is Discounted Reward (Sec. \ref{subsec:algorithm_ablation}), SR is Situation Removal (Sec. \ref{sec:reward_shaping}) with 0 reward and a physical reset upon progress reversal, and SPOT-Q is eq. \ref{eq:spotq}. 
% See the if statement in Alg. \ref{alg:spot_q_replay} for how SPOT-Q is applied during experience replay.
}
\vspace{-0.3cm}
\end{table}

\textbf{SPOT-Q:} VPG~\cite{zeng2018vpg} evaluated heuristics that specify exact locations to explore, and they found it led to \textit{worse} performance. 
A similar approach in QT-Opt~\cite{2018qtopt} is phased out as training proceeds, indicating that their methods do not contribute to improving outcomes throughout the training process. 
By contrast, SPOT-Q is enabled at all times and excises regions with zero likelihood of success, while other regions of interest remain open for exploration.
So does this difference in heuristic design matter?

The ``Mask but no SPOT-Q'' test disables the \texttt{if} statement in Alg. \ref{alg:spot_q_replay} to simulate a typical heuristic in which exploration is directed to particular regions without zero reward guidance.
% rtrial-no-mask-no-spotq: trial success (95+97+87+74)/4 = 88.25      efficiency (30+32+11+20)/4 = 23.25
% rtrial-mask-no-spot-q: trial success (95+99+92+93)/4 = 94.75 action efficiency (46+55+16+32)/4 = 37.25
% rtrial+spot-q: trial success (100+100+94+94)/4 = 97 efficiency (45+51+25+34)/4 = 38.75
% rprogress+spot-q: trial success (96+100+100+98)/4 = 98.5   efficiency (25+45+62+68)/4 = 50.0
``Mask but no SPOT-Q'' completes 95\% of trials, compared to 88\% without masking and 99\% with SPOT-Q;
action efficiency results are even more pronounced at 37\%, 23\%, and 50\% respectively. 
Both these results and Sec. \ref{sec:lava_gridworld_safety_domain_generalization} show SPOT-Q simply works throughout training and testing with little to no tuning, and so we conclude that SPOT-Q improves the efficiency of learning from heuristic data.
% and so we conclude that

\textbf{SPOT-Q Alternatives:} 
We evaluated two alternatives to SPOT-Q (eq. \ref{eq:spotq}, Alg. \ref{alg:spot_q_replay}), where 0 reward backpropagation is performed on all masked pixels with loss applied to the (1) sum, and (2) average of the masked scores in addition to the actually executed action.
In both cases, the gradients exploded and the algorithm did not converge. 
Only SPOT-Q is able to efficiently enhance convergence.

\textbf{Reward Weighting:} SPOT-Q + $\rprogress$ where $W_{push}\!=\!0.1$ succeeds in 99\% of trials, but just 27\% when $W_{push}\!=\!1.0$. 
The weighting in Fig. \ref{fig:trial-reward} on $\rtrial$ without masking or SPOT-Q achieves 97\% stack success and 38\% action efficiency, but we leave all weighting constant for consistency in Table \ref{table:StackingResults}.
This shows $W$ (eq. \ref{eq:vpg}) is important for efficient training.

\textbf{SPOT-Q + $\rprogress$:} This configuration has the best overall simulation performance with a 99\% trial success rate and 50\% efficiency, or about 10 actions per trial.
It is also the best simulated row model with 98\% trial success in one test and 100\% in the second, with a high 62-68\% action efficiency.
% rtrial-no-mask-no-spotq: trial success (95+97+87+74)/4 = 88.25      efficiency (30+32+11+20)/4 = 23.25
% rtrial-mask-no-spot-q: trial success (95+99+92+93)/4 = 94.75 action efficiency (46+55+16+32)/4 = 37.25
% rtrial+spot-q: trial success (100+100+94+94)/4 = 97 efficiency (45+51+25+34)/4 = 38.75
% rprogress+spot-q: trial success (96+100+100+98)/4 = 98.5   efficiency (25+45+62+68)/4 = 50.0

\textbf{SPOT-Q + $\rtrial$:} This has the best stacking model with 100\% completion in both test cases, and 45-51\% efficiency.
Overall performance is the second best with 97\% trial success, and 37\% efficiency, or about 14 actions per trial.
% , while still retaining the sparsity benefits of $\rdiscount$.

\begin{table}\centering
\vspace{0.4cm}
% \ra{1.3}
\setlength\tabcolsep{1.5pt} % default value: 6pt
\begin{tabular}{LCCCCCCC@{}}\toprule
\rowcolor{white}\textbf{Real} &\multicolumn{2}{c}{Domain} & Trials &  Action && & Training\\
\rowcolor{white}Test Task &Train& Test & Complete & Efficiency& Reward&SPOT-Q& Actions\\
\midrule
Clear 20 Toys & Real&Real &1/1& 75\% &$\rtrial$&\cmark& 1k\\ % costar 2020-02-23-18-51-58_Real-Push-and-Grasp-SPOT-Trial-Reward-Common-Sense-Testing
\midrule

\rowcolor{white}Stack of 4& Real&Real & 82\% & 60\% &$\rtrial$&\cmark&2.5k\\ % 2020-02-09-11-02-57_Real-Stack-SPOT-Trial-Reward-Common-Sense-Training
% \textbf{Stack of 4} &$\rtrial$& \textbf{Sim}&\textbf{Real} & \textbf{90\%} & \textbf{59\%}&\cmark&\textbf{10k}\\ % 2020-02-22-17-52-17_Real-Stack-SPOT-Trial-Reward-Common-Sense-Testing

% sim to real stack rprogress no-spotq, real test Testing Complete! Dir: /media/costar/f5f1f858-3666-4832-beea-b743127f1030/real_good_robot/logs/2020-06-05-19-42-28_Real-Stack-Two-Step-Reward-Testing Testing results: {'trial_success_rate_best_value': 1.0, 'trial_success_rate_best_index': 116, 'grasp_success_rate_best_value': 0.581081081081081, 'grasp_success_rate_best_index': 116, 'place_success_rate_best_value': 0.8837209302325582, 'place_success_rate_best_index': 116, 'action_efficiency_best_value': 0.5172413793103449, 'action_efficiency_best_index': 118}
Stack of 4& Sim&Real & 100\% & 51\% &$\rprogress$&\xmark&20k\\

% sim to real stack rprogress spot-q, real test     > Testing Complete! Dir: /media/costar/f5f1f858-3666-4832-beea-b743127f1030/real_good_robot/logs/2020-06-07-16-31-33_Real-Stack-Two-Step-Reward-Masked-Testing > Testing results: >  {'trial_success_rate_best_value': 1.0, 'trial_success_rate_best_index': 108, 'grasp_success_rate_best_value': 0.6666666666666666, 'grasp_success_rate_best_index': 109, 'place_success_rate_best_value': 0.9090909090909091, 'place_success_rate_best_index': 110, 'action_efficiency_best_value': 0.6111111111111112, 'action_efficiency_best_index': 110}
\rowcolor{white}\textbf{Stack of 4} & \textbf{Sim}&\textbf{Real} & \textbf{100\%} & \textbf{61\%}&$\rprogress$&\cmark&\textbf{20k}\\

% sim to real stack rtrial spot-q, real test 2020-06-05-18-28-46_Real-Stack-SPOT-Trial-Reward-Masked-Testing {'trial_success_rate_best_value': 1.0, 'trial_success_rate_best_index': 108, 'grasp_success_rate_best_value': 0.703125, 'grasp_success_rate_best_index': 108, 'place_success_rate_best_value': 0.8888888888888888, 'place_success_rate_best_index': 110, 'action_efficiency_best_value': 0.6111111111111112, 'action_efficiency_best_index': 110}
\textbf{Stack of 4} & \textbf{Sim}&\textbf{Real} & \textbf{100\%} & \textbf{61\%}&$\rtrial$&\cmark&\textbf{20k}\\

\midrule
% \rowcolor{white}\textbf{Row of 4} & \textbf{Sim}&\textbf{Real} & \textbf{80\%} & \textbf{71\%} &$\rtrial$& \cmark&\textbf{10k}\\ % 2020-02-23-00-36-39_Real-Rows-SPOT-Trial-Reward-Common-Sense-Testing % Our action efficiency was 107 at 6 actions per row, which is 71\% assuming it takes 4 actions per row 1.07*(4/6).

% sim to real row rprogress no spot-q Testing Complete! Dir: /media/costar/f5f1f858-3666-4832-beea-b743127f1030/real_good_robot/logs/2020-06-05-19-14-58_Real-Rows-Two-Step-Reward-Testing Testing results:  {'trial_success_rate_best_value': 0.9, 'trial_success_rate_best_index': 53, 'grasp_success_rate_best_value': 0.7741935483870968, 'grasp_success_rate_best_index': 54, 'place_success_rate_best_value': 0.7916666666666666, 'place_success_rate_best_index': 55, 'action_efficiency_best_value': 1.2452830188679245, 'action_efficiency_best_index': 55}
% Rows Efficiency conversion 6 actions ideal -> 4 actions ideal 124.5*4/6=83
\rowcolor{white}\rowcolor{white}Row of 4& Sim&Real & 90\% & 83\% &$\rprogress$& \xmark&20k\\

% sim to real row rprogress spot-q > Testing Complete! Dir: /media/costar/f5f1f858-3666-4832-beea-b743127f1030/real_good_robot/logs/2020-06-07-17-19-34_Real-Rows-Two-Step-Reward-Masked-Testing    > Testing results:    > {'trial_success_rate_best_value': 1.0, 'trial_success_rate_best_index': 82, 'grasp_success_rate_best_value': 0.68, 'grasp_success_rate_best_index': 83, 'place_success_rate_best_value': 0.8181818181818182, 'place_success_rate_best_index': 83, 'action_efficiency_best_value': 0.8780487804878049, 'action_efficiency_best_index': 84}
% Rows Efficiency conversion 6 actions ideal -> 4 actions ideal 87.8*4/6=58.5
\textbf{Row of 4} & \textbf{Sim}&\textbf{Real} & \textbf{100\%} & \textbf{59\%} &$\rprogress$& \cmark&\textbf{20k}\\
% TODO(ahundt) Note: the spot-q masked trial reward run might have been trained be back from when there were a few bugs affecting performance, so ideally we should redo training before doing this test, but we have limited access to the real robot due to COVID-19 so we ran with the sim model we had.

% sim to real row rtrial spot-q    > Testing Complete! Dir: /media/costar/f5f1f858-3666-4832-beea-b743127f1030/real_good_robot/logs/2020-06-07-18-10-48_Real-Rows-SPOT-Trial-Reward-Masked-Testing > Testing results: >  {'trial_success_rate_best_value': 0.9, 'trial_success_rate_best_index': 76, 'grasp_success_rate_best_value': 0.8333333333333334, 'grasp_success_rate_best_index': 76, 'place_success_rate_best_value': 0.7428571428571429, 'place_success_rate_best_index': 77, 'action_efficiency_best_value': 0.868421052631579, 'action_efficiency_best_index': 78}
% Rows Efficiency conversion 6 actions ideal -> 4 actions ideal 8786.8*4/6=57.8
\rowcolor{white}\textbf{Row of 4} & \textbf{Sim}&\textbf{Real} & \textbf{90\%} & \textbf{58\%} &$\rtrial$& \cmark&\textbf{20k}\\
\bottomrule
\end{tabular}
\caption{\label{table:RealResults}Real robot task results (Sec. \ref{subsec:real_robot_experiments}) with the SPOT framework. 
Bold entries highlight sim to real transfer with SPOT-Q. In this table no SPOT-Q also means no masking.
% TODO(ahundt) update video link, the video link below is a test done for the 2020-02 release, before several improvements got us to 100% test accuracy
% Video: \url{https://youtu.be/QHNkghXCmY0}
}
\vspace{-0.3cm}
\end{table}

\subsection{Safety and Domain Generalization}
\label{sec:lava_gridworld_safety_domain_generalization}
\begin{wrapfigure}{l}{0.28\columnwidth}
% \vspace{-0.2cm}
\vspace{0.1cm}
    \centering
    \includegraphics[width=0.25\columnwidth]{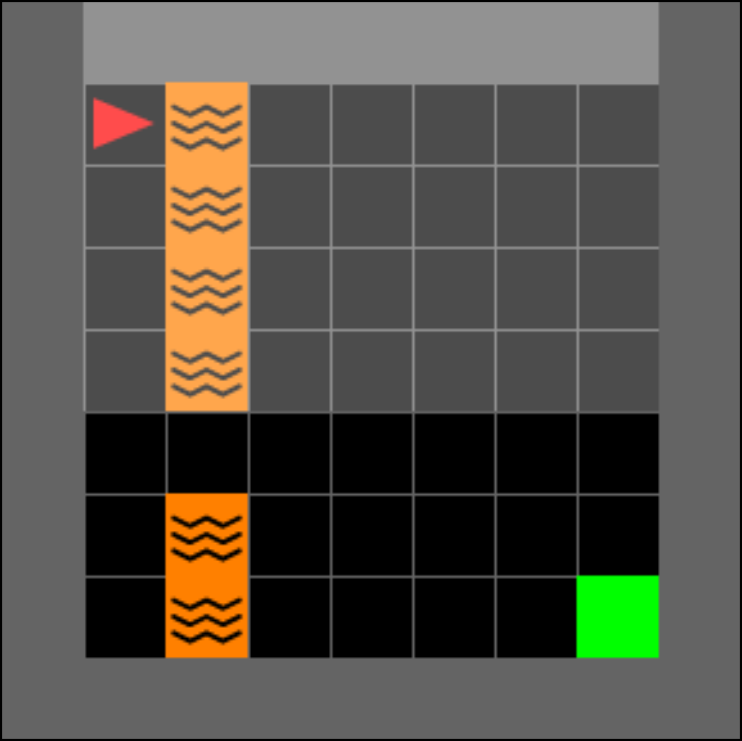}
    \caption{
    \label{fig:safety_gridworld}
    Safety Grid World where the goal is to avoid lava and get to the green square.}
% \vspace{-0.3cm}
\vspace{0.2cm}
\end{wrapfigure}
% Thus far we have focused on robotic tasks with superficial similarities in the scene and objects, so to 
To demonstrate the broad scope of the SPOT framework, we evaluate it on the simple but challenging Safety Grid World~\cite{gym_minigrid} (Fig. \ref{fig:safety_gridworld}), an environment type widely used to evaluate RL algorithms~\cite{ng1999policy,amodei2016safety}.
% $\mathtt{MiniGrid-LavaCrossingS9N1-v0}$.
Here the red robot must move forward or turn as it navigates towards the green square without ever entering the lava. 
If we had just one real robot to learn within this world, standard DRL would be extremely unsafe, but the SPOT framework 
allows the robot to safely explore the space.

As Table \ref{table:gridworld_results} shows, all improvements are consistent with our more realistic tasks.
We start with Rainbow~\cite{hessel2018rainbow}, a Q learning based DRL method, which only completes at most 12\% of trials within 500k actions with a 12\% efficiency.
We then perform a small ablation study, successively adding Masking, SPOT-Q, and $\rprogress$ to Rainbow; 96.9\%, 95.5\%, and 99.9\% of 1000 test trials are completed, respectively; average efficiency is 75\%, 73\%, and 62\%, respectively; and the average number of actions to complete 100\% of 30 validation trials is 123k, 113k, and 70k, respectively\footnote{In the grid world we only evaluate $\rprogress$ and the built-in reward (where all reward is delivered at the end) because there is little distinction between a failed action and failed trial.}. 
All failures with a mask did not enter the lava, they hit a 100 action limit.

These results are consistent with our more realistic experiments, demonstrate how the SPOT framework generalizes across completely different scenarios, and illustrate the application of the SPOT framework to safe exploration.
Next, we demonstrate how the SPOT framework leverages knowledge acquired in simulation directly on a real robot task.

% \begin{wraptable}{c}{0.5\columnwidth}\centering
\begin{table}
\vspace{0.4cm}
\centering
% \vspace{-0.2cm}
% \ra{1.3}
\setlength\tabcolsep{3pt} % default value: 6pt
\begin{tabular}{CCCCCCC@{}}\toprule
\rowcolor{white}Mask&SPOT-Q&$\rprogress$&Trials Complete& Efficiency & Actions\\
\midrule
\xmark&\xmark&\xmark&10.9-12.2\%&11-12\% & $>\!500$k\\
% first run 250k second run 150k
% mask only efficiency (62+80)/2 = 71, (120+330)/2 = 225.
\rowcolor{white}\cmark&\xmark&\xmark &93.9-96.9\%&62-80\%&100-145k\\
% mask spotq efficiency (76+89)/2 = 82.5, actions (125+340)/2 = 232.5
\cmark&\cmark&\xmark &95.0-96.0\%&72-74\% &100-125k\\
% mask + spotq + RP efficiency (86+89)/2 = 87.5 actions (125+305)/2 = 215
\rowcolor{white}\cmark&\cmark&\cmark &\textbf{99.8-99.9\%}&\textbf{62-64\%} &\textbf{65-75k}\\
\bottomrule
\end{tabular}

\caption{\label{table:gridworld_results}
Safety Grid World  (Fig. \ref{fig:safety_gridworld}) comparison of algorithm changes on top of Rainbow~\cite{hessel2018rainbow}. 
Cases without $\rprogress$ use the built-in reward. 
``Trials Complete'' is the percentage of 1000 test trials successfully completed by reaching the green square in fewer than 100 actions without entering lava.
``Efficiency'' is the test \texttt{Ideal/Actual} actions per trial after 500k training actions.
The ideal action count for each trial is found via a wavefront planner.
% The ``Reward'' column is the training time reward function.
% ``Test $R_{GW}$'' is the max over 1m actions, up to 0.96.
``Actions'' reports how many training steps were taken until the first case where 100\% of 30 validation trials succeed.
Values are the min and max of two runs.
}
% \vspace{-0.7cm}
\vspace{-0.3cm}
\end{table}
% \end{wraptable}

\section{Real World Experiments}
\label{subsec:real_robot_experiments}
%In this section, we demonstrate the translation of the SPOT-Q framework to the real robot. 
%We compare both training from scratch on the real robot and transfer of learned models from the simulation.
Finally, we examine the performance of SPOT-Q on real robot tasks, both via training from scratch and sim to real transfer. 
In both cases, performance was roughly equivalent to that achieved in simulation, which shows the strength of our approach for efficient and effective reinforcement learning.
%Both approaches lead to performance equivalent to that achieved in simulation, and there are few differences between training on real-world data and direct transfer of simulated models to the real world.
We use the setup described in \cite{paxton2018evaluating,hundt2019costar}, including a Universal Robot UR5, a Robotiq 2-finger gripper, and a Primesense Carmine RGB-D camera; all but the arm differ from those in our simulation.
Other implementation details are as described in Sec \ref{subsec:robot_implementation_details}, and results are in Table \ref{table:RealResults}. 

% \paragraph{Real Robot Training}

% === real robot training stuff goes here

% \textbf{Real Stacking:} We train from scratch on the real robot to perform the block stacking tasks .
% We will compare this real world baseline against running our simulation trained model on the real robot without fine tuning.

\textbf{Real Pushing and Grasping:} 
We train the baseline pushing and grasping task from scratch in the real world, test with 20 objects and see 100\% test clearance, 75\% grasp success rate, and 75\% efficiency in 1k actions; these results are comparable to the performance charted by VPG~\cite{zeng2018vpg} over 2.5k actions.
Sim to real does not succeed in this task.

\begin{figure}[bt!]
\vspace{0.1cm}
% \vspace{-0.5cm}
    \centering
    % \hfill
    % \includegraphics[width=\columnwidth]{grasp_place_stack_efficiency}
    % \includegraphics[width=\columnwidth]{2020-02-09-11-02-57_Real-Stack-SPOT-Trial-Reward-Common-Sense-Training-Real-Stack-SPOT-Trial-Reward-Common-Sense-Training_success_plot.png}
    \includegraphics[width=0.8\columnwidth]{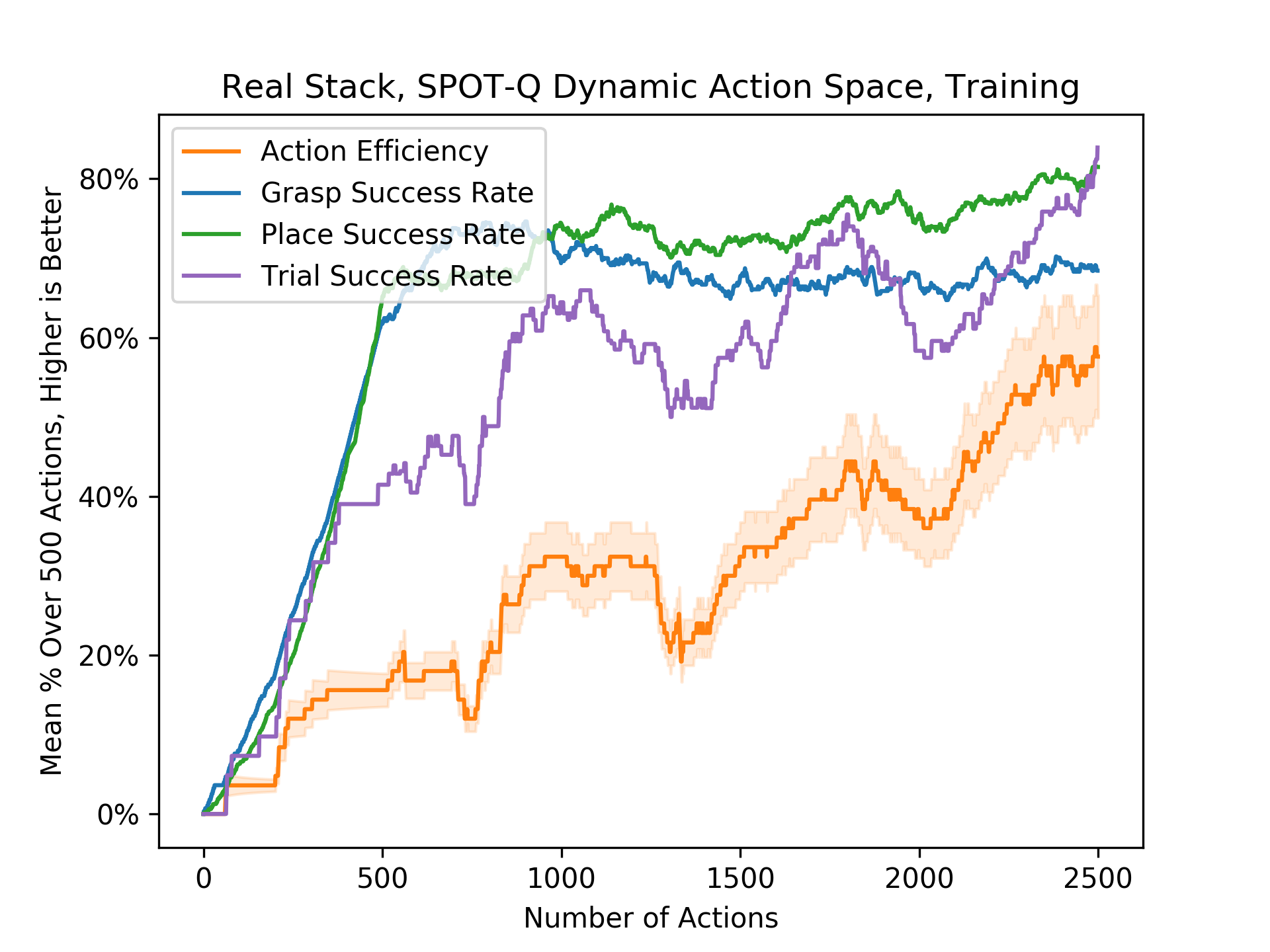}
    \caption{
    \label{fig:real_stack_training}
    Real training of the SPOT framework to Stack 4 Cubes with $\rtrial$ and SPOT-Q. 
    Failures include missed grasps, off-stack placements, and actions in which the stack topples. 
    Toppling can occur during successful grasp and push actions. 
}
\vspace{-0.3cm}
\end{figure}

\textbf{Sim to Real vs Real Stacking:} After training in simulation we directly load the model for execution on the real robot. 
Remarkably, all tested sim to real stacking models complete 100\% of trials, outperforming a model trained on the real robot which is successful in 82\% of trials (Fig.~\ref{fig:real_stack_training}, Table~\ref{table:RealResults}).
$\rprogress$ and $\rtrial$ have an equal action efficiency at 61\%, and the version of $\rprogress$ without SPOT-Q or a mask exhibits slightly lower efficiency at 51\%.
This is particularly impressive considering that our scene is exposed to variable sunlight.
Intuitively, these results are in part due to the depth heightmap input in stacking and row-making.

\textbf{Sim to Real Rows:} Our $\rprogress$ + SPOT-Q sim to real rows model is also able to create rows in a remarkable 100\% of attempts with 59\% efficiency.
$\rtrial$ + SPOT-Q and $\rprogress$ with no mask perform slightly worse, both with 90\% of trials complete, and an efficiency of 83\% and 58\%, respectively.
The high efficiency of $\rprogress$ with no mask is because we end real trials immediately when the task becomes unrecoverable, such as when a block tumbles out of the workspace.
We exclusively evaluate sim to real transfer in this case because training progress is significantly slower than with stacks.

We expect that block based tasks are able to transfer because the network relies primarily on the depth images, which are more consistent between simulated and real data.
This might reasonably explain why pushing and grasping does not transfer, a problem which could be mitigated in future work with methods like Domain Adaptation~\cite{tobin2017simtorealworld,bousmalis2018using}.

\section{Conclusion}

We have demonstrated that the SPOT framework is effective for training long-horizon tasks.
% , in particular for those where it is easy to reverse progress. 
To our knowledge, this is the first instance of reinforcement learning with successful sim to real transfer applied to long term multi-step tasks such as block-stacking and creating rows with consideration of \textit{progress reversal}.
% First, our EVT neural network model far exceeds existing methods' computational efficiency for manipulation tasks while simultaneously providing a 20\% increase in action efficiency and a 15\% higher perfect completion rate for adversarial pushing and grasping scenarios.
% Our results show the continued importance of neural network architecture design choices for Robotics and Reinforcement Learning algorithms. 
The SPOT framework quantifies an agent's progress within multi-step tasks while also providing zero-reward guidance, a masked action space, and situation removal. 
It is able to quickly learn policies that generalize from simulation to the real world. 
We find these methods are necessary to achieve a 100\% completion rate on both the real block stacking task and the row-making task. 

SPOT's main limitation is that while intermediate rewards can be sparse, they are still necessary. 
Future research should look at ways of learning task structures that incorporate situation removal from data.
In addition, the action space mask $M$ is currently manually designed; this and the lower-level open loop actions might be learned as well.
Another topic for investigation is the difference underlying successful sim to real transfer of stacking and row tasks when compared to pushing and grasping.
Finally, in the future, we would like to apply our method to more challenging tasks.

\section*{Acknowledgment}

This material is based upon work supported by the NSF NRI Awards \#1637949 and \#1763705. 
Hongtao Wu's contribution was funded under Office of Naval Research Award N00014-17-1-2124, Gregory S. Chirikjian, PI. 
We extend our thanks to Adit Murali for Safety Grid World integration; to Molly O'Brien for valuable discussions, feedback, and editing; to Corinne Hundt for the ``Good Robot!'' title copywriting;
to Michelle Hundt, Thomas Hundt, and Ian Harkins for editing;
to all those who gave their time for reading, reviewing, and feedback; 
and to the VPG~\cite{zeng2018vpg} authors for releasing their code.

\bibliographystyle{IEEEtran}
\bibliography{main}

% \appendix
% \input{10appendix.tex}

\end{document}